\documentclass[10pt,twocolumn,letterpaper]{article}

\usepackage{iccv}
\usepackage{times}
\usepackage{epsfig}
\usepackage{graphicx}
\usepackage{amsmath}
\usepackage{amssymb}
\usepackage{cite}
\usepackage[utf8]{inputenc} 
\usepackage[T1]{fontenc}    
\usepackage{url}            
\usepackage{booktabs}       
\usepackage{amsfonts}       
\usepackage{nicefrac}       
\usepackage{microtype}      
\usepackage{xcolor}
\usepackage{paralist}
\DeclareMathOperator*{\argmax}{argmax}

\definecolor{citecolor}{HTML}{0071bc}
\usepackage{multirow}
\usepackage[title]{appendix}
\usepackage{xspace}
\usepackage{float}
\makeatletter
\DeclareRobustCommand\onedot{\futurelet\@let@token\@onedot}
\def\@onedot{\ifx\@let@token.\else.\null\fi\xspace}
\def\eg{\emph{e.g}\onedot} 
\def\ie{\emph{i.e}\onedot}

\def\etal{\emph{et al}\onedot}
\makeatother

\usepackage[export]{adjustbox}
\definecolor{citecolor}{HTML}{0071bc}

\usepackage[pagebackref=true,breaklinks=true,letterpaper=true,colorlinks,citecolor=citecolor,bookmarks=false]{hyperref}

\iccvfinalcopy 

\def\iccvPaperID{***} 

\begin{document}

\title{On Equivariant and Invariant Learning of Object Landmark Representations}

\author{Zezhou Cheng \quad \quad Jong-Chyi Su \quad \quad  Subhransu Maji\\
University of Massachusetts Amherst\\
{\tt\small \{zezhoucheng, jcsu, smaji\}@cs.umass.edu}
}

\maketitle

\begin{abstract}
Given a collection of images, humans are able to discover landmarks by modeling the shared geometric structure across instances.
This idea of geometric equivariance has been widely used for the unsupervised discovery of object landmark representations.
In this paper, we develop a simple and effective approach by combining instance-discriminative and spatially-discriminative contrastive learning.
We show that when a deep network is trained to be invariant to geometric and photometric transformations, representations emerge from its intermediate layers that are highly predictive of object landmarks.
Stacking these across layers in a ``hypercolumn'' and projecting them using spatially-contrastive learning further improves their performance on matching and few-shot landmark regression tasks.
We also present a unified view of existing equivariant and invariant representation learning approaches through the lens of contrastive learning, shedding light on the nature of invariances learned.
Experiments on standard benchmarks for landmark learning, as well as a new challenging one we propose, show that the proposed approach surpasses prior state-of-the-art. 
\end{abstract}

\section{Introduction}
\label{Introduction}
Learning in the absence of labels is a challenge for existing machine learning and computer vision systems. 
Despite recent advances, the performance of unsupervised learning remains far below that of supervised learning, especially for few-shot image understanding tasks.
This paper considers the task of unsupervised learning of object landmarks from a collection of images.
The goal is to learn representations that can be used to establish correspondences across objects, and to predict landmarks such as eyes and noses when provided with a few labeled examples.

\begin{figure}
\centering
\includegraphics[width=0.95\linewidth]{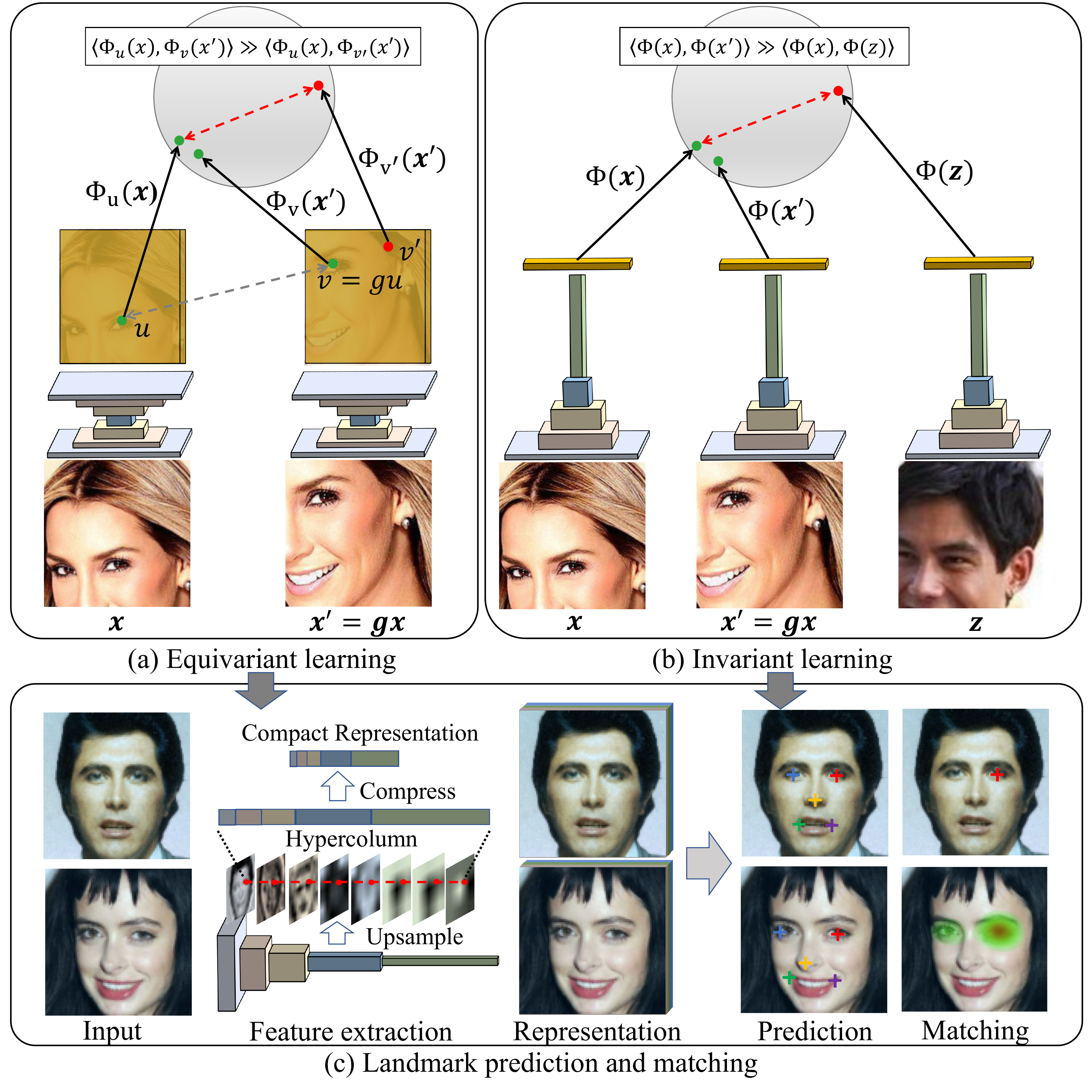}
\caption{\textbf{Equivariant and invariant learning.} \textbf{(a)} Equivariant learning requires representations across locations to be invariant to a geometric transformation $g$ while being distinctive across locations. \textbf{(b)} Invariant learning encourages the representations to be invariant to transformations while being distinctive across images. Thus both can be seen as instances of contrastive learning. \textbf{(c)} A hypercolumn feature and its compact representation are highly predictive of object landmarks.}
\label{fig:splash}
\vspace{-0.2in}
\end{figure}

One way of inferring structure is to reason about the global appearance in terms of disentangled factors such as geometry and texture.
This is the basis of alignment based~\cite{miller2000learning,huang2012learning} and generative modeling based approaches for landmark discovery~\cite{zhang2018unsupervised,wiles2018self,shu2018deforming,jakab2018unsupervised,jakabself,xu2020unsupervised}.
An alternate is to learn a representation that geometrically transforms in the same way as the object, a property called \emph{geometric equivariance} (Fig.~\ref{fig:splash}a)~\cite{thewlis2017factorized,thewlis2017dense,DVE}.
However, useful invariances may not be learned (\eg, the raw pixel representation itself is equivariant), limiting their applicability in the presence of clutter, occlusion, and inter-image variations.

A different line of work has proposed instance discriminative \emph{contrastive learning} as an unsupervised objective~\cite{hadsell2006dimensionality,dosovitskiy2014discriminative, hjelm2018learning,bachman2019learning,Wu2018a,Tian2019,he2019momentum,chen2020simple,oord2018representation,zhuang2019local,henaff2019data}.
The goal is to learn a representation $\Phi$ that has higher similarity between an image $\mathbf{x}$ and its transformation $\mathbf{x}'$ than with a different one $\mathbf{z}$, \ie, $\langle \Phi(\mathbf{x}), \Phi(\mathbf{x}')\rangle \gg \langle \Phi(\mathbf{x}), \Phi(\mathbf{z})\rangle$, as illustrated in Fig.~\ref{fig:splash}b.
A combination of geometric (\eg, cropping and scaling) and photometric (\eg, color jittering and blurring) transformations are used to encourage the representation to be \emph{invariant} to these transformations while being \emph{distinctive} across images.
Recent work~\cite{he2019momentum,chen2020simple,chen2020improved,chen2020big} has shown that contrastive learning is effective, even outperforming ImageNet~\cite{deng2009imagenet} pre-training on various tasks.
However, to predict landmarks a representation cannot be invariant to geometric transformations. 
This paper asks the question: \emph{are equivariant losses necessary for unsupervised landmark discovery?}
In particular, do representations predictive of object landmarks automatically emerge in intermediate layers of a deep network trained to be invariant to image transformations? 
While empirical evidence suggests that semantic parts emerge when deep networks are trained on supervised tasks~\cite{zhou2016learning,gonzalez2018semantic}, is it also the case for unsupervised learning? 

This work aims to address these by presenting a unified view of the equivariant and invariant learning approaches.
We show that when a deep network is trained to be invariant to geometric and photometric transformations, its intermediate-layer representations are highly predictive of landmarks (Fig.~\ref{fig:splash}b).
The emergence of invariance and the loss of geometric equivariance is gradual in the representation hierarchy, a phenomenon that has been studied empirically~\cite{zeiler2014visualizing,lenc2015understanding} and theoretically~\cite{tishby2000information, tishby2015deep,achille2018emergence}.
This observation motivates a \emph{hypercolumn} representation~\cite{hariharan2015hypercolumns}, which we find to be more effective for landmark predictions (Fig.~\ref{fig:splash}c).

We also observe that objectives used in equivariant learning can be seen as a contrastive loss between representations across locations within the \emph{same image}, as opposed to invariant learning where the loss is applied \emph{across images}~(Fig.~\ref{fig:splash}).
This observation sheds light on the nature of the invariances learned by the two approaches.
It also allows us to obtain a compact representation of the high-dimensional hypercolumns simply by learning a linear projection under the spatially contrastive objective.
The projection results in spatially distinctive representations and significantly improves the landmark matching performance (Tab.~\ref{tab:matching} and Fig.~\ref{fig:matching}).

To validate these claims, we perform experiments by training deep networks using Momentum Contrast (MoCo)~\cite{he2019momentum} on several landmark matching and detection benchmarks.
Other than commonly used ones, we also present a comparison by learning on a challenging dataset of birds from the iNaturalist dataset~\cite{gvanhorn2018inat} and evaluating on the CUB dataset~\cite{wah2011caltech}.
We show that the contrastive-learned representations (without supervised regression) can be predictive in landmark matching experiments. 
For landmark detection, we adapt the commonly used linear evaluation setting by varying the number of labeled examples (Fig.~\ref{fig:landmarks} \& \ref{fig:curves}).
Our approach is simple, yet it offers consistent improvements over prior approaches~\cite{thewlis2017factorized,thewlis2017dense,zhang2018unsupervised,jakab2018unsupervised,DVE} (Tab.~\ref{table: face regression}).
While the hypercolumn representation leads to a larger embedding dimension, it comes at a modest cost as our approach outperforms the prior state-of-the-art~\cite{DVE}, with as few as 50 annotated training examples on the AFLW benchmark~\cite{koestinger2011annotated} (Fig.~\ref{fig:curves}).
Furthermore, we use dimensionality reduction based on the equivariant learning to improve the performance on landmark matching (Tab.~\ref{tab:matching}), as well as landmark prediction in the low data regime (Tab.~\ref{tab:low-d}).

\section{Related Work}
\label{sec:related}

\noindent
\textbf{Background.} A representation $\Phi: {\cal X} \rightarrow \mathbb{R}^C$ is said to be equivariant (or covariant) with a
transformation $g$ for input $\mathbf{x} \in {\cal X}$ if there exists a map $M_g : \mathbb{R}^C \rightarrow \mathbb{R}^C$  such that: $\forall \mathbf{x} \in {\cal X}: \Phi(g\mathbf{x}) \approx M_g\Phi(\mathbf{x})$. In other words, the representation transforms in a predictable manner given the input transformation.
For natural images, the transformations can be geometric (\eg, translation, scaling, and rotation), photometric (\eg, color changes), or more complex (\eg, occlusion, viewpoint or instance variations).
Note that a sufficient condition for equivariance is when $\Phi$ is invertible since $M_g = \Phi^{-1}\circ g$.
Invariance is a special case of equivariance when $M_g$ is the identity function, \ie, $\phi(g\mathbf{x}) \approx \phi(\mathbf{x})$. There is a rich history in computer vision on the design of covariant (\eg, SIFT~\cite{lowe2004distinctive}), and invariant representations (\eg, HOG~\cite{dalal2005histograms} and Bag-of-Visual-Words~\cite{sivic2003video}).

\noindent
\textbf{Deep representations.} Invariance and equivariance in deep network representations result from both the architecture (\eg, convolutions lead to translational equivariance, while pooling leads to translational invariance), and learning (\eg, invariance to categorical variations).
Lenc \etal~\cite{lenc2015understanding} showed that early-layer representations of a deep network are nearly equivariant as they can be ``inverted'' to recover the input, while later layers are more invariant.
Similar observations have been made by visualizing these representations~\cite{mahendran2016visualizing,zeiler2014visualizing}.
The gradual emergence of invariance can also be theoretically understood in terms of a ``information bottleneck'' in the feed-forward hierarchy~\cite{achille2018emergence,tishby2000information,tishby2015deep}.
While equivariance to geometric transformations is relevant for landmark representations, the notion can be generalized to other transformation groups~\cite{gens2014deep,cohen2016group}.

\noindent
\textbf{Landmark discovery.}
Empirical evidence~\cite{zhou2016learning,oquab2015object} suggests that semantic parts emerge when deep networks are trained on supervised tasks.
This has inspired architectures for image classification that encourage part-based reasoning, such as those based on texture representations~\cite{lin2015bilinear,cimpoi2015deep,arandjelovic2016netvlad} or spatial attention~\cite{sermanet2014attention,xiao2015application,fu2017look}.
In contrast, our work shows that \emph{parts also emerge when models are trained in an unsupervised manner.}
When no labels are available, equivariance to geometric transformations provides a natural self-supervisory signal. 
The equivariance constraint requires $\Phi_u(\mathbf{x})$, the representation of $\mathbf{x}$ at location $u$, to be invariant to the geometric transformation $g$ of the image, \ie, $ \forall \mathbf{x}, u: \Phi_{gu}(g\mathbf{x}) = \Phi_u(\mathbf{x})$ (Fig.~\ref{fig:splash}a).
This alone is not sufficient since both $\Phi_u(\mathbf{x}) = \mathbf{x}_u$ and $\Phi_u(\mathbf{x}) = \text{constant}$ satisfy this property.
Constraints based on locality~\cite{thewlis2017dense,DVE} and diversity~\cite{thewlis2017factorized} have been proposed to avoid this pathology.
Yet, inter-image invariance is not directly enforced.
Another line of work is based on a generative modeling approach~\cite{zhang2018unsupervised,wiles2018self,shu2018deforming,jakab2018unsupervised,lorenz2019unsupervised,jakabself,sanchez2019object,xu2020unsupervised,bespalov2020brul}. These methods implicitly incorporate equivariant constraints by modeling objects as deformation (or flow) of a shape template together with appearance variation in a disentangled manner.
In contrast, our work shows that learning representations invariant to both geometric and photometric transformations is an effective strategy. 
These invariances emerge at different rates in the representation hierarchy, and can be selected with a small amount of supervision for the downstream task. 

\noindent
\textbf{Unsupervised learning.} Recent work has shown that unsupervised objectives based on density modeling~\cite{dosovitskiy2014discriminative, hjelm2018learning,oord2018representation,bachman2019learning,Wu2018a,Tian2019,he2019momentum,chen2020simple,chen2020improved,chen2020big} 
outperform unsupervised (or self-supervised) learning based on pretext tasks such as colorization~\cite{Zhang2016}, rotation prediction~\cite{Gidaris2018}, jigsaw puzzle~\cite{Noroozi2016}, and inpainting~\cite{Pathak2016}.
These contrastive learning objectives~\cite{hadsell2006dimensionality} are often expressed in terms of noise-contrastive estimation (NCE)~\cite{gutmann2010noise} (or maximizing mutual information~\cite{oord2018representation,hjelm2018learning}) between different views obtained by geometrically and photometrically transforming an image.
The learned representations thus encode invariances to these transformations while preserving information relevant for downstream tasks.
However, the effectiveness of unsupervised learning depends on how well these invariances relate to those desired for end tasks.
Despite recent advances, existing methods for unsupervised learning significantly lack in comparison to their supervised counterparts in the few-shot setting~\cite{Goyal2019}.
Moreover, their effectiveness for landmark discovery has not been sufficiently studied in the literature.\footnote{Note that MoCo~\cite{he2019momentum} was evaluated on pose estimation, however, their method was trained with 150K labeled examples and the entire network was fine-tuned.}
In part, it is not clear why invariance to geometric transformations might be useful for landmark prediction since we require the representation to carry some spatial information about the image.
Understanding these trade-offs and improving the effectiveness of contrastive learning for landmark prediction is one of the goals of the paper.
\section{Method}
\label{sec:method}
Let $\mathbf{x} \in \mathbb{R}^{H \times W \times 3}$ denote an image of an object, and $u \in \Omega = \{0, \dots, H-1\}\times\{0, \dots, W-1\}$ denote pixel coordinates. 
The goal is to learn a function $\Phi_u(\mathbf{x}): \Omega \rightarrow \mathbb{R}^C$ that outputs a pixel representation at spatial location $u$ of input $\mathbf{x}$ that is predictive for object landmarks. 
We assume $C \gg 3$ aiming to learn a high-dimensional representation of landmarks.
This is similar to~\cite{DVE} which learns a local descriptor for each landmark, and unlike those that represent them as a discrete set~\cite{zhang2015learning}, or on a planar ($C=2$)~\cite{thewlis2017factorized,zhang2018unsupervised} or spherical ($C=3$)~\cite{thewlis2017dense} coordinate system.
In other words the representation should be predictive of landmarks or effective for matching, without requiring compactness or topology in the embedding space.
Note that this is in contrast to some work on literature where a fixed set of landmarks are discovered (\eg,~\cite{thewlis2017factorized,zhang2018unsupervised,jakab2018unsupervised}). One may obtain this, for instance, by clustering the landmark representations in the embedding space. 

We describe commonly used equivariance constraints for unsupervised landmark discovery~\cite{thewlis2017factorized, thewlis2017dense, DVE}, followed by models based on invariant learning~\cite{oord2018representation,he2019momentum}.
We then present our approach that integrates the equivariant and invariant learning approaches.

\subsection{Equivariant and invariant representations}
\noindent
\textbf{Equivariant learning.}\label{sec:equi}
The equivariance constraint requires $\Phi_u(\mathbf{x})$, the representation of $\mathbf{x}$ at location $u$, to be invariant to the geometric deformation of the image (Fig.~\ref{fig:splash}a). Given a geometric warping function $g:\Omega \rightarrow \Omega$, the representation of $\mathbf{x}$ at $u$ should be same as the representation of the transformed image $\mathbf{x}'=g\mathbf{x}$ at $v=gu$, that is, $\forall \mathbf{x}, u\in \Omega: \Phi_{v}(\mathbf{x}') = \Phi_u(\mathbf{x})$. 
This constraint can be captured by the loss: 
\begin{equation}
\label{eq:equivariance}
    \mathcal{L}_{\mathit{equi}} = \frac{1}{|\Omega|} \sum_{u \in \Omega} \|\Phi_u(\mathbf{x}) -  \Phi_v(\mathbf{x}')\|^2.
\end{equation}
A diversity (or locality) constraint is necessary to encourage the representation to be distinctive across locations. 
For example, Thewlis~\etal~\cite{thewlis2017dense} proposed the following:
\begin{equation}
\label{eq:diversity}
    \mathcal{L}_{\mathit{div}} = \frac{1}{|\Omega|} \sum_{u \in \Omega} \|gu -  \argmax_v \langle \Phi_u(\mathbf{x}), \Phi_v(\mathbf{x}') \rangle\|^2, 
\end{equation}
which they replaced by a probabilistic version that combines both the losses as:
\begin{equation}
\label{eq:equivariance2}
    \mathcal{L}_{\mathit{equi}}' = \frac{1}{|\Omega|^2} \sum_{u \in \Omega}\sum_{v \in \Omega}\|gu-v\| ~p(v | u; \Phi, \mathbf{x}, \mathbf{x}').
\end{equation}
Here $p(v | u; \Phi, \mathbf{x}, \mathbf{x}')$ is the probability of pixel $u$ in image $\mathbf{x}$ matching $v$ in image $\mathbf{x}'$ with $\Phi$ as the encoder shared by $\mathbf{x}$ and $\mathbf{x}'$ computed as below, and $\tau \in \mathbb{R}^+$ is a scale parameter:
\begin{equation}
\label{eq:softdiv}
    p(v |u; \Phi, \mathbf{x}, \mathbf{x'}) = \frac{\text{exp}(\langle \Phi_u(\mathbf{x}), \Phi_v(\mathbf{x}') \rangle/\tau)}{\sum_{t \in \Omega} \text{exp}(\langle \Phi_u(\mathbf{x}), \Phi_t(\mathbf{x}') \rangle/\tau)}.
\end{equation}

\noindent
\textbf{Invariant learning.}\label{sec:invariance} Contrastive learning is based on the similarity over pairs of inputs (Fig.~\ref{fig:splash}b).
Given an image $\mathbf{x}$ and its transformation $\mathbf{x}'$ as well as other images $\mathbf{z}_i$, $i \in \{1,2...N\}$, the InfoNCE~\cite{oord2018representation} loss minimizes:
\begin{equation}
\label{eq:infonce}
\mathcal{L}_{\mathit{inv}} = -\log\frac{\exp\left(\langle\Phi(\mathbf{x}),\Phi(\mathbf{x}')\rangle\right)}{\sum_{i=1}^N\exp(\langle\Phi(\mathbf{x}),\Phi(\mathbf{z}_i)\rangle)}.
\end{equation}
The objective encourages representations to be invariant to transformations while being distinctive across images.
To address the computational bottleneck in evaluating the denominator, Momentum Contrast (MoCo)~\cite{he2019momentum} computes the loss over negative examples using a dictionary queue and updates the parameters based on momentum.

\noindent
\textbf{Transformations.} The space of transformations used to generate image pairs $(\mathbf{x}, \mathbf{x}')$ plays an important role in learning.
A common approach is to apply a combination of \emph{geometric transformations}, such as cropping, resizing, and thin-plate spline warping, as well as \emph{photometric transformations}, such as color jittering and adding JPEG noise.
Transformations can also denote channels of an image or modalities such as depth and color~\cite{Tian2019}.

\noindent
\textbf{Hypercolumns.} A deep network of $n$ layers (or blocks\footnote{Due to skip-connections, we cannot decompose the encoding over layers, but can across blocks.}) can be written as $\Phi(\mathbf{x}) = \Phi^{(n)} \circ \Phi^{(n-1)} \circ\dots\circ \Phi^{(1)}(\mathbf{x})$.
A representation $\Phi(\mathbf{x})$ of size $H' \times W' \times C$ can be spatially interpolated to the input size $H\times W \times C$ to produce a pixel representation $\Phi_u(\mathbf{x}) \in \mathbb{R}^C$.
The hypercolumn representation of layers $k_1, k_2, \dots, k_n$ is obtained by concatenating the interpolated features from the corresponding layers, that is,  $\Phi_u(\mathbf{x}) = \Phi^{(k_1)}_u(\mathbf{x}) \oplus \Phi^{(k_2)}_u(\mathbf{x})\oplus \dots \oplus \Phi^{(k_n)}_u(\mathbf{x})$.

\subsection{Approach}
Given a large unlabeled dataset, we first train representations using instance-discriminative contrastive learning framework of MoCo~\cite{he2019momentum}. A combination of geometric and photometric transformations are applied to generate pairs $(\mathbf{x}, \mathbf{x}')$. We then extract single layer or hypercolumn representations from the trained network to represent landmarks (Fig.~\ref{fig:splash}c). Subsequently, we incorporate a spatial contrastive learning to reduce dimensionality and induce spatial diversity by training a linear projector over the frozen landmark representation.
Let $w \in \mathbb{R}^{C \times d}$, where $d \ll C$, used to project the landmark representation as $\Phi'_u(x) = w^T\Phi_u(x)$.
The goal that the projected embeddings are spatially distinct within the same image, \ie, 
\begin{equation}
\label{eq:equivariance3}
    \forall u,v\in \Omega: u \neq v \Leftrightarrow \Phi'_u(\mathbf{x}) \neq \Phi'_v(\mathbf{x}), 
\end{equation}
is obtained by optimizing objective in Eqn.~\ref{eq:equivariance2} with $\mathbf{x}' = \mathbf{x}$.

\noindent
\textbf{Discussion.} 
Note that since the linear projection is location-wise, spatial equivariance is preserved but intra-image contrast is improved.
The projected embeddings are equally effective as the hypercolumn representations for landmark regression, but are significantly better for landmark matching (Tab.~\ref{tab:matching}). 
The intuition is that the hypercolumn features contain sufficient information about landmarks, but the projection step makes them spatially distinct which is more suitable for matching.
Novotny \etal~\cite{novotny2017anchornet} proposed a similar approach to extract compact representations for cross-instance semantic matching from a network pre-trained with class labels. In comparison, we only use unsupervised representations.
The idea of spatially contrastive learning has also been shown to be effective for learning scene-level representations~\cite{pinheiro2020unsupervised}.

\section{Experiments}
\label{sec:exp}

\begin{figure*} [ht]
    \setlength{\tabcolsep}{0.5pt}
    \newcommand\wifig{0.077\linewidth}
    \centering
     \begin{tabular}{c c c c c @{\hspace{3pt}} | @{\hspace{3pt}} cccc @{\hspace{3pt}} | @{\hspace{3pt}} cccc @{\hspace{3pt}}}
    \rotatebox[origin=c]{90}{Reference} &
    \includegraphics[width=\wifig, valign=c]{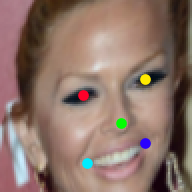} &
    \includegraphics[width=\wifig, valign=c]{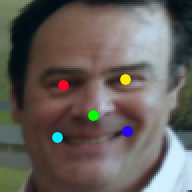} &
    \includegraphics[width=\wifig, valign=c]{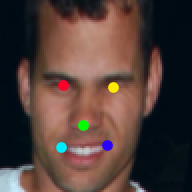} &
    \includegraphics[width=\wifig, valign=c]{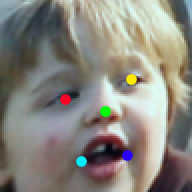} &
    \includegraphics[width=\wifig, valign=c]{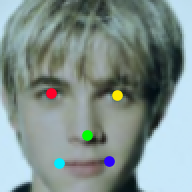} &
    \includegraphics[width=\wifig, valign=c]{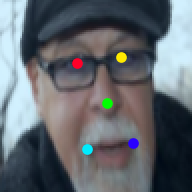} &
    \includegraphics[width=\wifig, valign=c]{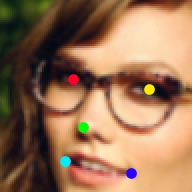} &
    \includegraphics[width=\wifig, valign=c]{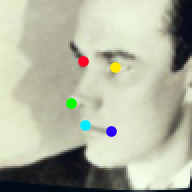} &
    \includegraphics[width=\wifig, valign=c]{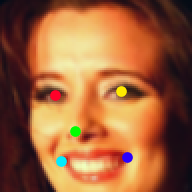} &
    \includegraphics[width=\wifig, valign=c]{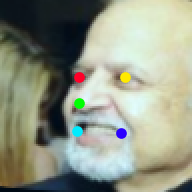} &
    \includegraphics[width=\wifig, valign=c]{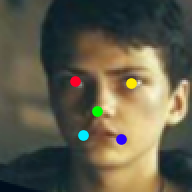} &
    \includegraphics[width=\wifig, valign=c]{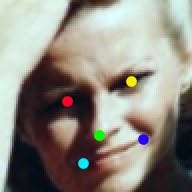} \\
        \rotatebox[origin=c]{90}{3840-D} &
    \includegraphics[width=\wifig, valign=c]{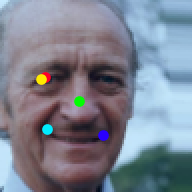} &
    \includegraphics[width=\wifig, valign=c]{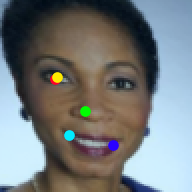} &
    \includegraphics[width=\wifig, valign=c]{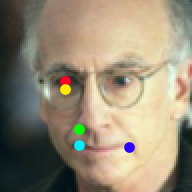} &
    \includegraphics[width=\wifig, valign=c]{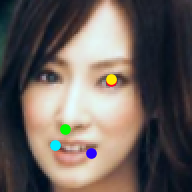} &
    \includegraphics[width=\wifig, valign=c]{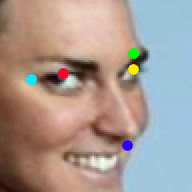} &
    \includegraphics[width=\wifig, valign=c]{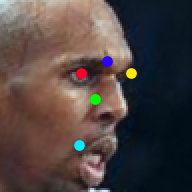} &
    \includegraphics[width=\wifig, valign=c]{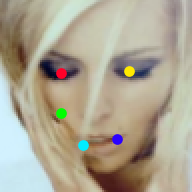} &
    \includegraphics[width=\wifig, valign=c]{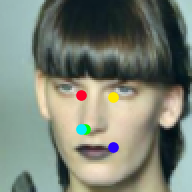} &
    \includegraphics[width=\wifig, valign=c]{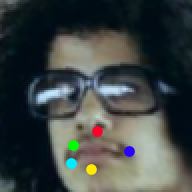} &
    \includegraphics[width=\wifig, valign=c]{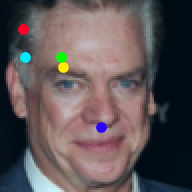} &
    \includegraphics[width=\wifig, valign=c]{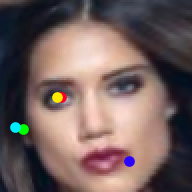} &
    \includegraphics[width=\wifig, valign=c]{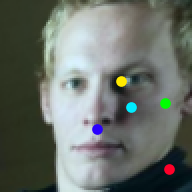} \\
        \rotatebox[origin=c]{90}{256-D} &
    \includegraphics[width=\wifig, valign=c]{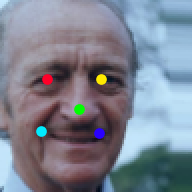} &
    \includegraphics[width=\wifig, valign=c]{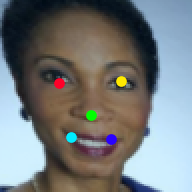} &
    \includegraphics[width=\wifig, valign=c]{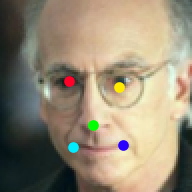} &
    \includegraphics[width=\wifig, valign=c]{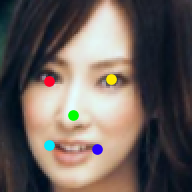} &
    \includegraphics[width=\wifig, valign=c]{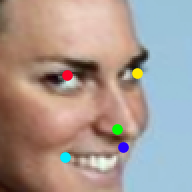} &
    \includegraphics[width=\wifig, valign=c]{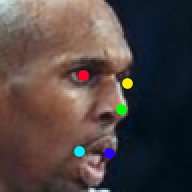} &
    \includegraphics[width=\wifig, valign=c]{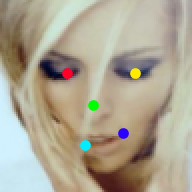} &
    \includegraphics[width=\wifig, valign=c]{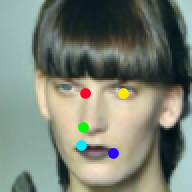} &
    \includegraphics[width=\wifig, valign=c]{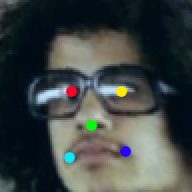} &
    \includegraphics[width=\wifig, valign=c]{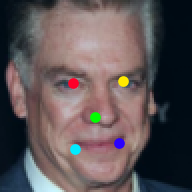} &
    \includegraphics[width=\wifig, valign=c]{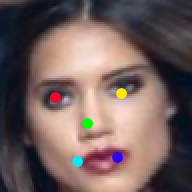} &
    \includegraphics[width=\wifig, valign=c]{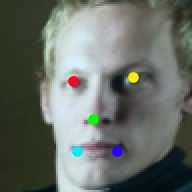} \\
    \end{tabular}
    \caption{\textbf{Visualization of landmark matching} with cosine distance using 3840-D hypercolumn features and 256-D features projected from hypercolumn. Failure cases of using hypercolumn includes (Left) mismatching between two eyes and (Middle) lack of robustness to large viewpoint or (Right) appearance changes across different identities. The proposed feature projection method alleviates these issues.}
    \label{fig:matching}
    \vspace{-7pt}
\end{figure*}

We first outline the datasets and implementation details of the proposed method (\S~\ref{sec:datasets}). We then evaluate our model and provide comparisons to the existing methods qualitatively and quantitatively on landmark matching (\S~\ref{sec:matching}) and landmark detection benchmarks (\S~\ref{sec:regression}). We conclude with ablation studies and discussions (\S~\ref{sec:discussion}).

\subsection{Benchmarks and implementation details} 
\label{sec:datasets}

\noindent
\textbf{Human faces.} We first compare the proposed model with prior art on the existing human face landmark detection benchmarks.
 Following DVE~\cite{DVE}, we train our model on aligned CelebA dataset~\cite{liu2015deep} and evaluate on MAFL~\cite{zhang2015learning}, AFLW~\cite{koestinger2011annotated}, and 300W~\cite{300W}. 
 The overlapping images with MAFL are excluded from CelebA.
 MAFL comprises 19,000 training images and 1000 test images with annotations on 5 face landmarks. 
 Two versions of AFLW are used: AFLW$_M$ which contains 10,122 training images and 2995 testing images, which are crops from MTFL~\cite{zhang2014facial}; AFLW$_R$ which contains tighter crops of face images with 10,122 for training and 2991 for testing. 
 300W provides 68 annotated face landmarks with 3148 training images and 689 test images.
 We apply the same image pre-processing procedures as in DVE, the current state-of-the-art, for a direct comparison.
 We also train our model on the unaligned raw CelebA dataset to evaluate the efficiency of representation learning on in-the-wild unlabeled images.

\noindent
\textbf{Birds.} We collect a challenging dataset of birds where objects appear in clutter, occlusion, and exhibit wider pose variation.
We randomly select 100K images of birds from the iNaturalist 2017 dataset~\cite{gvanhorn2018inat} under the ``Aves'' class to train unsupervised representations. For the performance in the few-shot setting, we collect a subset of CUB dataset~\cite{wah2011caltech} 
containing 35 species of \textit{Passeroidea}\footnote{This is the biggest Aves taxa in iNaturalist.} super-family, each annotated with 15 landmarks.
We sample at most 60 images per class which results in 1241 images as our training set, 382 as validation set, and 383 as test set 
(see appendix for the details). 

\noindent
\textbf{Evaluation.} We use landmark matching and detection as the end tasks for evaluation. 
In landmark matching, following DVE~\cite{DVE}, we generate 1000 pairs of images from the MAFL test set as the benchmark, among which 500 are pairs of the same identity obtained by warping images with thin-plate spline (TPS) deformation, and others are pairs of different identities. 
Each pair of images consists of a reference image with landmark annotations and a target image. 
We use the nearest neighbor matching with cosine distance between pixel representations for landmark matching, and report the mean pixel error between the predicted landmarks and the ground-truth landmarks.

\definecolor{mygray}{gray}{0.4}
\newcommand\grey[1]{\textcolor{mygray}{\textsuperscript{#1}}}
\begin{table}[t!]
    \centering
    \begin{tabular}{c c c c c c c}
        \toprule
        \multirow{2}{*}{Method} &  \multirow{2}{*}{Dim.} & \multicolumn{2}{c}{Aligned} & \multicolumn{2}{c}{In-the-wild} \\
            &  & Same & Diff. & Same & Diff.\\
        \midrule
        DVE  & 64 & 0.92 & 2.38 & 1.27 & 3.52 \\
        Ours  & 3840 & \textbf{0.73} & 6.16  & \textbf{0.78} & 5.58\\
        Ours + proj.  & 256 & \textbf{0.71} & \textbf{2.06} & \textbf{0.96} & \textbf{3.03} \\
        Ours + proj.  & 128 & \textbf{0.82} & \textbf{2.19} & \textbf{0.98} & \textbf{3.05} \\
        Ours + proj.  & 64 & \textbf{0.92} & 2.62  & \textbf{0.99} & \textbf{3.06} \\
        \bottomrule
    \end{tabular}
    \caption{\textbf{Landmark matching results.} We report the mean pixel error between the predicted landmarks and the ground-truth across 1000 pairs of images from MAFL (\emph{lower is better}). The test set consists of 500 same-identity and 500 different-identity pairs. We compare DVE~\cite{DVE} with Hourglass net and our models with ResNet50 trained from aligned or in-the-wild CelebA dataset. We also evaluate the effect of feature projection (+proj.) with different output dimensions. Our results better than DVE's~\cite{DVE} are marked in \textbf{bold}.} 
    \label{tab:matching}
    \vspace{-8pt}
\end{table}

In the landmark regression task, following~\cite{thewlis2017dense, DVE}, we train a linear regressor to map the representations to landmark annotations while keeping the representations frozen.
The landmark regressor is a linear regressor per target landmark. Each regressor consists of $K$ filters of size 1$\times$1$\times C$ on top of a $C$-dimensional representation to generate $K$ intermediate heatmaps, which are then converted to spatial coordinates by \texttt{soft-argmax} operation. These $K$ coordinates are finally converted to the target landmark by a linear layer (see appendix for details). 
We use $K=50$ to keep the evaluation consistent with prior works~\cite{thewlis2017dense, DVE}, but we find that this hyperparameter is not critical (see Sec.~\ref{sec:discussion}).
We report errors in the percentage of inter-ocular distance on face benchmarks and the percentage of correct keypoints (PCK) on CUB. 
A prediction is considered correct according to the PCK metric if its distance to the ground-truth is within 5\% of the longer side of the image.
The occluded landmarks are ignored during evaluation. 
We did not find fine-tuning to be uniformly beneficial but include a comparison in the appendix. 

\noindent
\textbf{Implementation details.}  We use MoCo~\cite{he2019momentum} to train our models on CelebA or iNat Aves for 800 epochs with a batch size of 256 and a dictionary size of 4096.
ResNet18 or ResNet50~\cite{he2016deep} are used as our backbones. 
We extract hypercolumns~\cite{hariharan2015hypercolumns} per pixel by stacking activations from the second (\texttt{conv2\_x}) to the last convolutional block (\texttt{conv5\_x}). 
We resize the feature maps from the selected convolutional blocks to the same spatial size as DVE~\cite{DVE} (\ie~48$\times$48). 
We also follow DVE (with Hourglass network) to resize the input image to $136 \times 136$ then center-crop the image to $96 \times 96$ for face datasets. 
Images are resized to $96 \times 96$ without any cropping on the bird dataset.
For a comparison with DVE on the CUB dataset we used their publicly available implementation. 
More details are in the appendix.

\subsection{Landmark matching}
\label{sec:matching}

\noindent
\textbf{Quantitative results.} Tab.~\ref{tab:matching} compares the proposed method with DVE~\cite{DVE} quantitatively. 
We train DVE and our models on both aligned and in-the-wild unaligned version of CelebA dataset, and report the mean pixel error on aligned face images from MAFL. 
Our hypercolumn representation has high performance in same-identity matching but is not robust to cross-identity variations.
However, the proposed feature projection makes the hypercolumn more suitable for landmark matching. 
We experiment with different feature dimensions after projection and find that our method with 128 or higher dimensional features achieves the state-of-art. 
DVE outperforms ours with 64-D features when the representations are learned on the aligned CelebA dataset. 
This is because the architecture of the Hourglass network and the joint training of the backbone and feature extractor enables DVE to learn a more compact representation than our method.
However, to lift the feature dimension from 64 to 256, DVE requires re-training the entire model while we only need to re-train a linear feature projector. 
Moreover, when the representation is learned from the in-the-wild CelebA dataset, our model \emph{outperforms DVE by a large margin}.
This suggests our representation is more invariant to nuisance factors than that of DVE. 
We also observe that our method with smaller networks (\eg~ResNet18) with 128-D projected features outperforms DVE, and both DVE and our methods outperform representations from ImageNet pretrained networks. (see appendix for more details).

\begin{table*}[h!]
  \setlength{\tabcolsep}{3.5pt}
  \centering
  \begin{tabular}{l c c @{\hskip 0.2in} c c c c@{\hskip 0.2in} c}
    \toprule
    \multirow{2}{*}{Method}   & \# Params.  & Unsuper.   & MAFL & AFLW$_M$ & AFLW$_R$ & 300W & CUB \\
     & Millions  &   & \multicolumn{4}{c}{Inter-ocular Distance (\%) $\downarrow$} & PCK $\uparrow$ \\
    \midrule
    TCDCN \cite{zhang2015learning}  & --   & $\times$ & 7.95 & 7.65 & -- & 5.54 & --\\
    RAR    \cite{xiao2016robust}    & -- & $\times$ &  --  & 7.23 & -- & 4.94 & -- \\
    MTCNN   \cite{zhang2014facial}  & --  & $\times$ & 5.39 & 6.90 & -- & -- & -- \\
    Wing Loss \cite{feng2018wing}   & -- & $\times$ &  --  & -- & -- & 4.04  & --\\
    \midrule
    \multicolumn{2}{l}{\textbf{Generative modeling based}}\\
    Structural Repr. \cite{zhang2018unsupervised} & -- & \checkmark & 3.15 & -- & 6.58 & -- & --\\
    FAb-Net \cite{wiles2018self}    & -- & \checkmark & 3.44 & -- & -- & 5.71& -- \\
    Deforming AE \cite{shu2018deforming} & --   & \checkmark & 5.45 & -- & -- & -- & --\\
    ImGen. \cite{jakab2018unsupervised} & --     & \checkmark & 2.54 & -- & 6.31 & -- & -- \\
    ImGen.++ \cite{jakabself} & -- & \checkmark & -- & -- & -- & 5.12  & -- \\
    \midrule
    \textbf{Equivariance based}\\
    Sparse \cite{thewlis2017factorized} & --    & \checkmark & 6.67 & 10.53 & -- & 7.97 & --\\
    Dense 3D \cite{thewlis2017dense} & --    & \checkmark & 4.02 & 10.99 & 10.14 & 8.23 & --\\
    DVE SmallNet \cite{DVE} & 0.35 & \checkmark & 3.42 & 8.60 & 7.79 & 5.75 & -- \\
    DVE Hourglass \cite{DVE} & 12.61 & \checkmark & 2.86 & 7.53 & 6.54 & 4.65  &  61.91 \\
    \midrule
   \textbf{Invariance based} \\
    Ours (ResNet18) & 11.24  & \checkmark & \textbf{2.57} & 8.59 & 7.38 & 5.78  & \textbf{62.24}\\
    Ours (ResNet18 + proj.)          &  11.24  &  \checkmark & \textbf{2.71} & \textbf{7.23} & \textbf{6.30} & 5.20 & 58.49 \\
    Ours (ResNet50)  & 23.77 & \checkmark & \textbf{2.44} & \textbf{6.99}  & \textbf{6.27}  & 5.22 & \textbf{68.63} \\
    Ours (ResNet50 + proj.)  & 23.77 & \checkmark & \textbf{2.64} & \textbf{7.17} & \textbf{6.14} & 4.99 & \textbf{62.55} \\ 
    \bottomrule
  \end{tabular}
   \caption{\textbf{Results on landmark detection.} Comparison on face benchmarks, including MAFL, AFLW$_M$, AFLW$_R$, and 300W, and CUB dataset. We report the error in the percentage of inter-ocular distance on human face dataset (\emph{lower is better}), and the percentage of correct keypoints (PCK) on CUB dataset (\emph{higher is better}). 
  We project the hypercolumn (\ie + proj.) to 256-D features on face and 512-D on bird dataset and provide results of other dimensions in the appendix.
  Our results better than DVE's~\cite{DVE} are marked in \textbf{bold}.}
  \label{table: face regression}
  \vspace{-9pt}
\end{table*}

\noindent\textbf{Qualitative results.} Fig.~\ref{fig:matching} presents the qualitative results of landmark matching. Our method with hypercolumn for matching is not robust to viewpoint and appearance changes, and frequently mismatches the left and right eyes. Incorporating the proposed feature projection adds diversity effectively and solves these issues. 

\subsection{Landmark detection} \label{sec:regression}

\noindent\textbf{Quantitative results.} 
Tab.~\ref{table: face regression} presents a quantitative evaluation of multiple benchmarks. 
On faces, our model with a ResNet50 achieves state-of-the-art results on all benchmarks except for 300W. 
On iNat Aves $\rightarrow$ CUB, out approach outperforms prior state-of-the-art~\cite{DVE} by a large margin, suggesting improved invariance to nuisance factors.
Incorporating the feature projection results in small performance degradation in some cases but remains the state-of-art. 
Our method with ResNet18 is comparable with DVE and benefits from using a deeper network.
We present more results of landmark detection under different configurations in the appendix.

\begin{figure} [htpb]
    \setlength{\tabcolsep}{1pt}
    \newcommand\wfig{0.142\linewidth}
    \centering
    \begin{tabular}{ccc @{\hspace{8pt}} rcccc}
    \rotatebox[origin=c]{90}{MAFL} &
    \includegraphics[width=\wfig, valign=c]{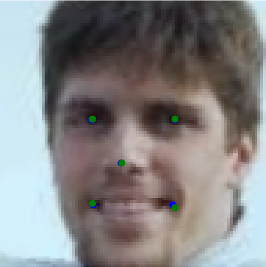} &
    \includegraphics[width=\wfig, valign=c]{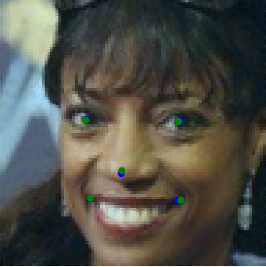} &
    \rotatebox[origin=c]{90}{GT} &
    \includegraphics[width=\wfig, valign=c]{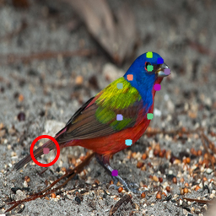} &
    \includegraphics[width=\wfig, valign=c]{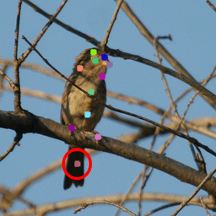} &
    \includegraphics[width=\wfig, valign=c]{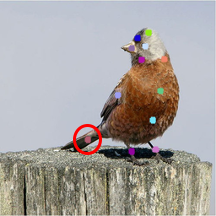} &
    \includegraphics[width=\wfig, valign=c]{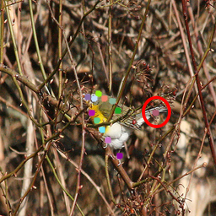}\\
    \rotatebox[origin=c]{90}{AFLW} &
    \includegraphics[width=\wfig, valign=c]{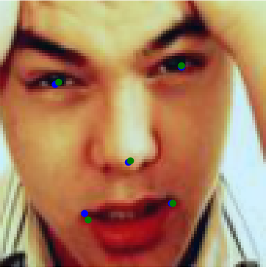} &
    \includegraphics[width=\wfig, valign=c]{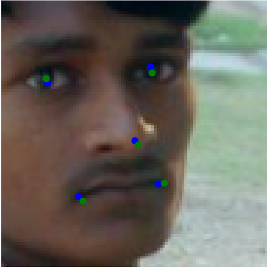} &
    \rotatebox[origin=c]{90}{DVE} &
    \includegraphics[width=\wfig, valign=c]{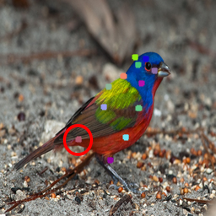} &
    \includegraphics[width=\wfig, valign=c]{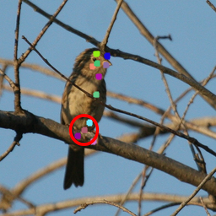} &
    \includegraphics[width=\wfig, valign=c]{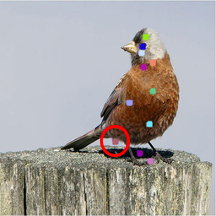} &
    \includegraphics[width=\wfig, valign=c]{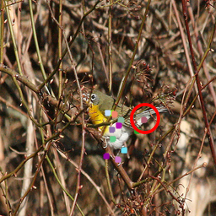}\\
    \rotatebox[origin=c]{90}{300W} &
    \includegraphics[width=\wfig, valign=c]{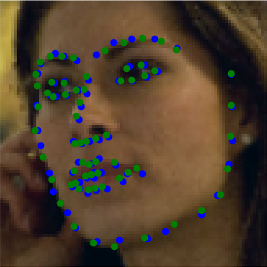} &
    \includegraphics[width=\wfig, valign=c]{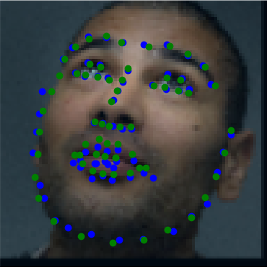} &
    \rotatebox[origin=c]{90}{Ours} &
    \includegraphics[width=\wfig, valign=c]{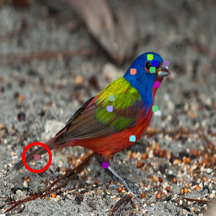} &
    \includegraphics[width=\wfig, valign=c]{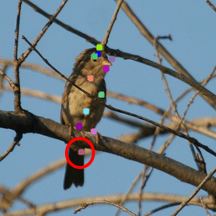} &
    \includegraphics[width=\wfig, valign=c]{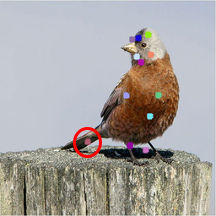} &
     \includegraphics[width=\wfig, valign=c]{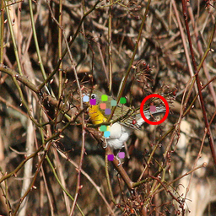}\\
    & \multicolumn{2}{c}{(a) Human face} & \multicolumn{5}{c}{(b) Birds} \\
    \end{tabular}
    \caption{\textbf{Detected landmarks} \textbf{(a)} on faces (\emph{blue:} predictions, \emph{green:} ground truth) and \textbf{(b)} on CUB. Notice that our method localizes tails of birds (circled) much better. \emph{Zoom in for details.}}
    \vspace{-10pt} 
    \label{fig:landmarks}
\end{figure}

\noindent\textbf{Qualitative results.} Fig.~\ref{fig:landmarks} shows qualitative results of landmark regression on human faces and birds. We notice that both DVE and our model with hypercolumn representations are able to localize the foreground object accurately. However, our model localizes many keypoints better (\eg, on the tails of the birds) and is more robust to the background clutter (\eg, the last column of Fig.~\ref{fig:landmarks}b).

\begin{figure*}[ht!]
    \setlength{\tabcolsep}{0.1pt}
    \centering
    \begin{tabular}{ccc}
    \includegraphics[width=0.31\linewidth]{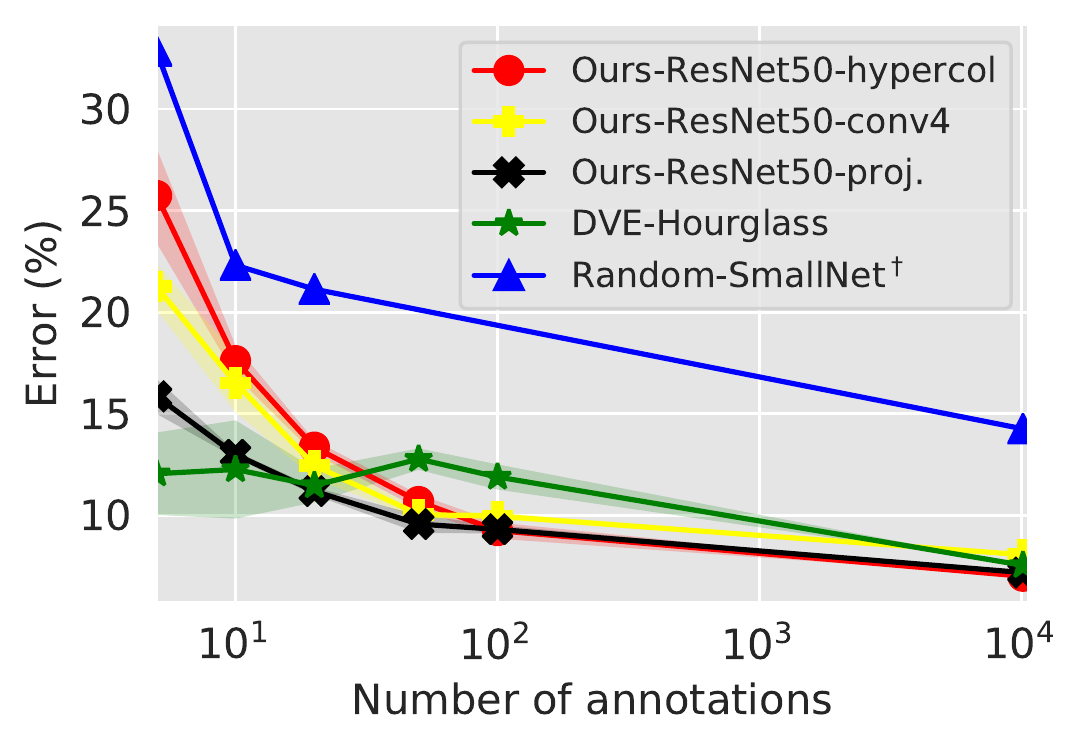} &
    \includegraphics[width=0.31\linewidth]{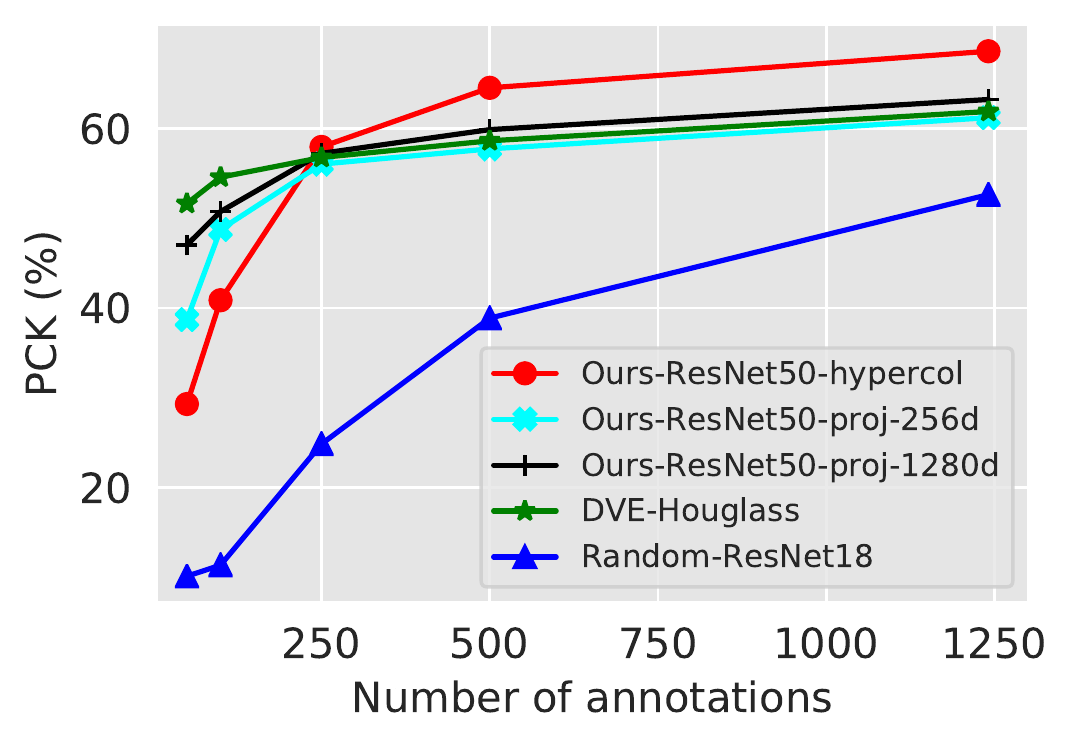} &
    \includegraphics[width=0.31\linewidth]{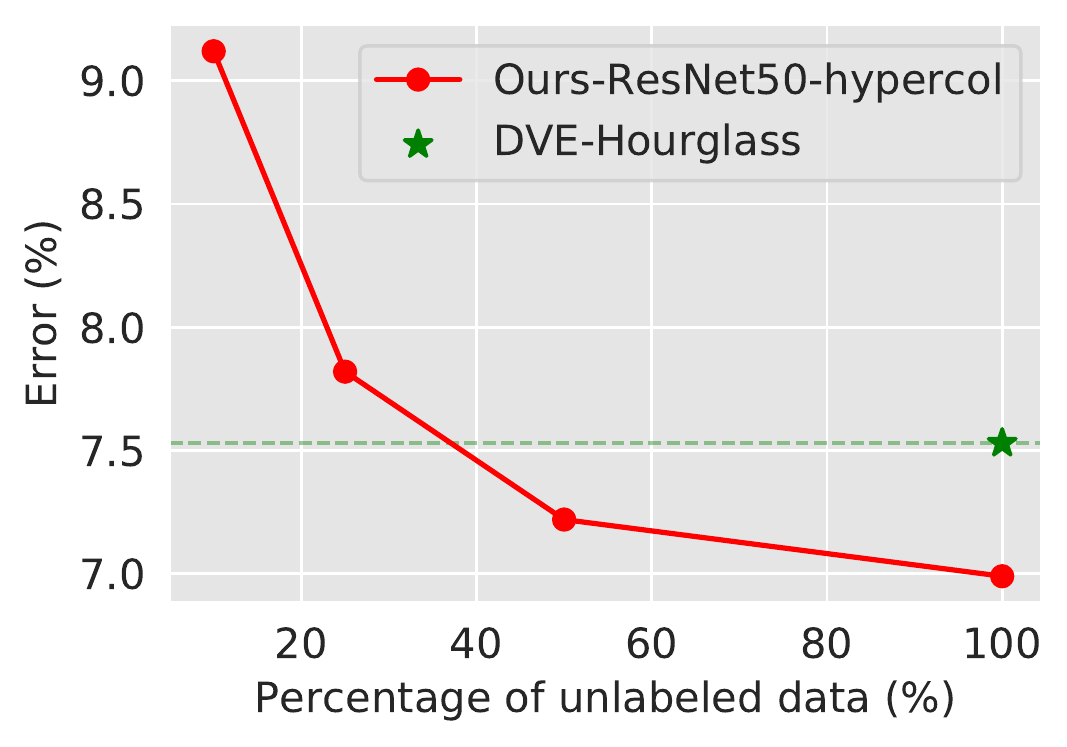} \\
    (a) Limited anno.~on AFLW$_M$ & (b) Limited anno.~on CUB  & (c) Unlabeled CelebA images
    \end{tabular}
    \caption{\textbf{The effect of dataset size.} 
    \textbf{(a)} A comparison of our model with DVE~\cite{DVE} by varying the number of annotations for landmark regression on AFLW$_M$ dataset. \texttt{Random-SmallNet}$^\dagger$: is a randomly initialized ``small network'' taken from~\cite{DVE}. \texttt{Ours-ResNet50}: are based on hypercolumn, or its compact representations, or fourth-layer features trained using contrastive learning.
    \textbf{(b)} Similar results on CUB dataset. \texttt{Random-ResNet18}: is trained from scratch on the CUB dataset.
    \textbf{(c)} Results of landmark regression on AFLW$_M$ using different numbers of \emph{unlabeled} images from CelebA for training.}
    \label{fig:curves}
    \vspace{-7pt} 
\end{figure*}

\noindent
\textbf{Limited annotations.} Fig.~\ref{fig:curves}a and \ref{fig:curves}b compare our model with DVE~\cite{DVE} using a limited number of annotations on AFLW$_M$ and CUB dataset respectively. 
Without feature projection, our performance is better as soon as a few training examples are available (\eg, 50 on AFLW$_M$ and 250 on CUB). 
This can be attributed to the higher dimensional embedding of the hypercolumn representation. 
The scheme can be improved by using a single-layer representation as shown in the yellow line. 
Our feature projection further improves the performance in the low-data regime as shown in the black line. Interestingly, this improvement is not solely due to the dimension reduction: increasing the dimension of projected feature from 256 to 1280 improves the performance across different dataset sizes on CUB (see Fig.~\ref{fig:curves}b).
Note that all unsupervised learning models (including DVE and our model) outperform the randomly initialized baseline on both human face and bird datasets. Numbers corresponding to Fig.~\ref{fig:curves} are in the appendix.

\begin{table*}[htbp!]
  \setlength{\tabcolsep}{5pt}
  \centering
  \begin{tabular}{c | c c c c c | c c c c c}
    \toprule
     \multirow{3}{*}{Dataset} & \multicolumn{5}{c}{Single layer} & \multicolumn{4}{|c}{Hypercolumn}         \\
        & \#1 & \#2 & \#3 & \#4 & \#5 &  \#4~-~\#5 &  \#3~-~\#5  &  \#2~-~\#5&   \#1~-~\#5  \\
            & (64)  & (256) & (512) & (1024) & (2048) & (3072) & (3584) & (3840) & (3904) \\
    \midrule
    MAFL         & 5.77        &    4.58       &  3.03      &   \textbf{2.73}    &    3.66        &   2.73       &  2.65     &  \textbf{2.44}   &    2.51       \\
    AFLW$_M$     & 24.20      &    21.34     &  11.95     &   \textbf{8.83}    &    11.55      &   8.14       & 8.31       & \textbf{6.99}   &  7.40         \\
    AFLW$_R$     &  16.27     &    14.15     &  9.66     &  \textbf{7.37}   &    8.83        &    6.95      &   \textbf{6.24}    &  6.27   &   6.34      \\
    300W         &  16.45     &    13.08     &  7.66     &   \textbf{6.01}   &    7.70        &   5.68       &  5.28     &  5.22   &    \textbf{5.21}      \\
    \bottomrule
  \end{tabular}
   \caption{\textbf{Landmark detection using single layer and hypercolumn representations.} The error is reported in the percentage of inter-ocular distance using linear regression over individual layers (left) and combinations (right), with a ResNet50. 
  The embedding dimension for each is shown in parenthesis. Layer \#4 performs the best across datasets, while hypercolumns offer an improvement.}
  \label{table: embedding}
  \vspace{-9pt} 
\end{table*}

\noindent
\textbf{Limited unlabeled data.} Fig.~\ref{fig:curves}c shows that our model with hypercolumn representation matches the performance of DVE on AFLW$_M$ using only 40\% of the images on the CelebA dataset. This suggests that invariances are acquired more efficiently in our framework.

\subsection{Ablation studies and discussions}\label{sec:discussion}
\noindent
\textbf{Hypercolumns.} Tab.~\ref{table: embedding} compares the performance of using individual layer and hypercolumn representations. The activations from the fourth convolutional block consistently outperforms those from the other layers.
For an input of size 96$\times$96, the spatial dimension of the representation is 48$\times$ 48 at Layer \#1 and 3$\times$3 at Layer \#5, reducing by a factor of two at each successive layer. 
Thus, while the representation loses geometric equivariance with depth, contrastive learning encourages invariance, resulting in Layer \#4 with the optimal trade-off for this task.
While the best layer can be selected with some labeled validation data, the hypercolumn representation provides further benefits everywhere except the very small data regime (Tab.~\ref{table: embedding} and Fig.~\ref{fig:curves}a).

\noindent
\textbf{Dimensionality and linear regressor.} In Tab.~\ref{tab:low-d}, 
we reduce the size of the landmark regressor to evaluate its effect on the landmark regression performance.
We chose 50 intermediate landmarks to keep the evaluation consistent with DVE. However, the choice is not critical as seen by the performance of a smaller linear regressor.
There is a small drop in performance, while it remains comparable to DVE. 
The proposed feature projection with equivariant learning is more effective than non-negative matrix factorization (NMF), a classical dimension reduction method.
\begin{table}[H]
    \centering
        \setlength{\tabcolsep}{1.4pt}
    \begin{tabular}{c c c c c c c c}
        \toprule
        Method & $C$ & $K$ & \#P & MAFL & AFLW$_M$ & AFLW$_R$ & 300W \\
        \midrule
        DVE & 64 & 50 & 17 & 2.86 & 7.53 & 6.54 & 4.65\\
        \hline
        Ours         &3840 & 50 & 961 & \textbf{2.44} & \textbf{6.99} & \textbf{6.27} & 5.22  \\
        Ours         &3840 & 10 & 192 & \textbf{2.40} & \textbf{7.27} & \textbf{6.30} & 5.40 \\
        Ours+proj. & 256 & 50 & 65 &  \textbf{2.64} & \textbf{7.17} &  \textbf{6.14} & 4.99 \\
        Ours+proj. & 256 & 10 & 13 & \textbf{2.67} & \textbf{7.24} & \textbf{6.23} & 5.07 \\
        Ours+proj. & 64  & 50 & 17  & \textbf{2.77} & \textbf{7.21} & \textbf{6.22} & 5.19\\
        Ours+NMF & 64 & 50  & 17 & \textbf{2.80} & 7.60 & 6.69 & 5.62 \\
        \bottomrule
    \end{tabular}
    \caption{\textbf{The effect of landmark regressor on landmark regression}. We vary the number of parameters (\#P in thousands) in the landmark regressor by changing the number of intermediate landmarks ($K$) and feature dimensions ($C$). We compare the proposed feature projection (\ie +proj.) with non-negative matrix factorization (NMF) for dimension reduction. Our results better than DVE's~\cite{DVE} are marked in \textbf{bold}.}
    \label{tab:low-d}
    \vspace{-5pt} 
\end{table}

\noindent
\textbf{Effectiveness of unsupervised learning.} Tab.~\ref{table: random net} compares representations using the linear evaluation setting for randomly initialized, ImageNet pretrained, and contrastively learned networks using a hypercolumn representation.
Contrastive learning provides significant improvements over ImageNet pretrained models, which is less surprising since the domain of ImageNet images is quite different from faces. 
Interestingly, random networks have competitive performances with respect to some prior work in Tab.~\ref{table: face regression}. 
For example, \cite{thewlis2017dense} achieve 4.02\% on MAFL, while a randomly initialized ResNet18 with hypercolumns achieves 4.00\%.

\begin{table}[ht!]
   \setlength{\tabcolsep}{2.6pt}
     \centering
  \begin{tabular}{c c c c c c c c}
    \toprule
    Network & Supervision  & MAFL & AFLW$_M$ & AFLW$_R$ & 300W \\
    \midrule
    \multirow{3}{*}{Res.~18} & Random          & 4.00 & 14.20 & 10.11 & 9.88  \\
    & ImageNet    & 2.85 & 8.76 & 7.03 & 6.66 \\
    & Contrastive   & 2.57 & 8.59 & 7.38 & 5.78 \\
    \midrule
    \multirow{3}{*}{Res.~50} & Random    & 4.72 & 16.74 & 11.23 & 11.70 \\
    & ImageNet  & 2.98 & 8.88 & 7.34 & 6.88 \\
    & Contrastive        & \textbf{2.44} & \textbf{6.99} & \textbf{6.27} & \textbf{5.22}\\
    \bottomrule
  \end{tabular}
   \caption{\label{table: random net}
    \textbf{Effectiveness of unsupervised learning.} Error using randomly initialized, ImageNet pretrained, and contrastively trained ResNet50 for landmark detection. Frozen hypercolumn representations with a linear regression were used for all methods. The results with feature projection are included in the appendix.}
    \vspace{-2pt} 
\end{table}

\begin{figure} [t!]
    \setlength{\tabcolsep}{0.5pt}
    \newcommand\wifig{0.101\linewidth}
    \centering
    \begin{tabular}{ccccccc @{\hspace{2pt}} | @{\hspace{2pt}} ccc }
    \rotatebox[origin=c]{90}{Part1} &
    \includegraphics[width=\wifig, valign=c]{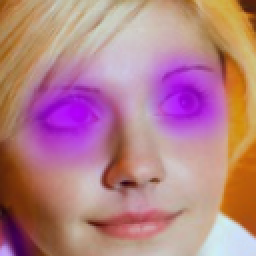} &
    \includegraphics[width=\wifig, valign=c]{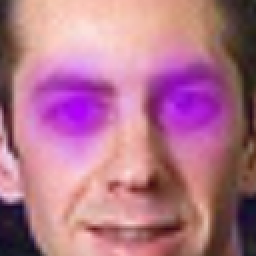} &
    \includegraphics[width=\wifig, valign=c]{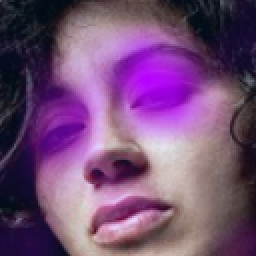} &
    \includegraphics[width=\wifig, valign=c]{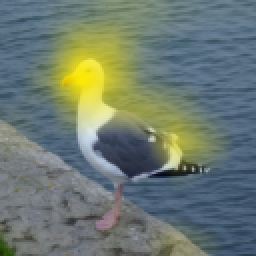} &
    \includegraphics[width=\wifig, valign=c]{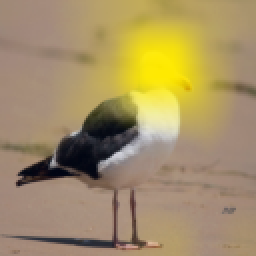} &
    \includegraphics[width=\wifig, valign=c]{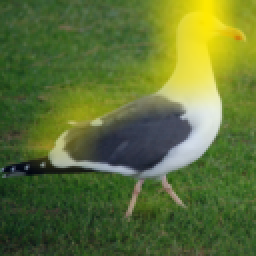} &
    \includegraphics[width=\wifig, valign=c]{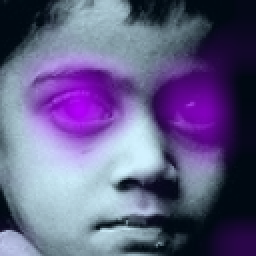} &
    \includegraphics[width=\wifig, valign=c]{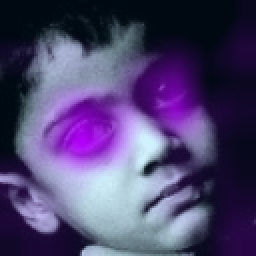} &
    \includegraphics[width=\wifig, valign=c]{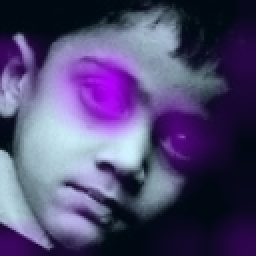} \\
    \rotatebox[origin=c]{90}{Part2} &
    \includegraphics[width=\wifig, valign=c]{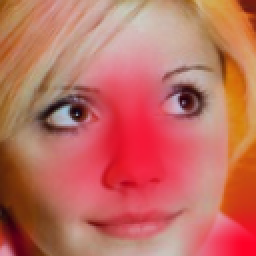} &
    \includegraphics[width=\wifig, valign=c]{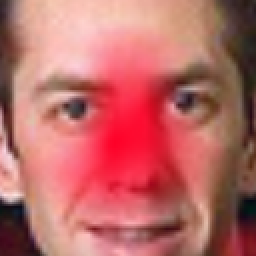} &
    \includegraphics[width=\wifig, valign=c]{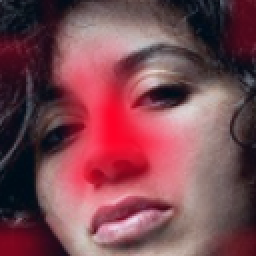} &
    \includegraphics[width=\wifig, valign=c]{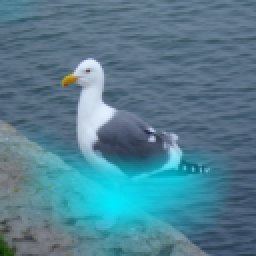} &
    \includegraphics[width=\wifig, valign=c]{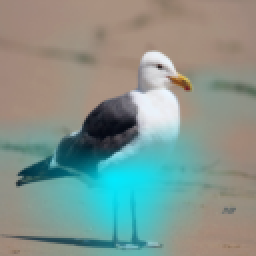} &
    \includegraphics[width=\wifig, valign=c]{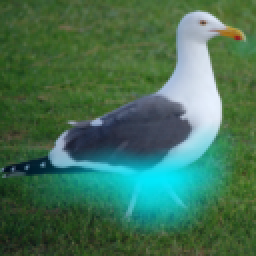} &
    \includegraphics[width=\wifig, valign=c]{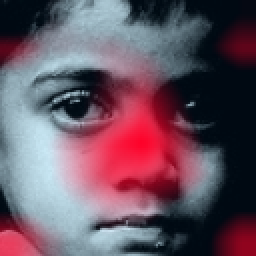} &
    \includegraphics[width=\wifig, valign=c]{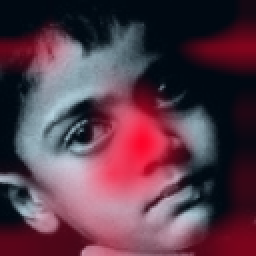} &
    \includegraphics[width=\wifig, valign=c]{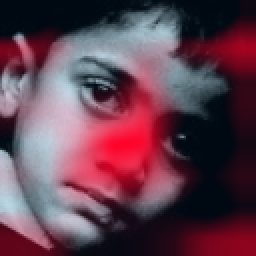} \\
    \rotatebox[origin=c]{90}{Map} &
    \includegraphics[width=\wifig, valign=c]{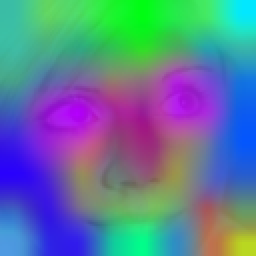} &
    \includegraphics[width=\wifig, valign=c]{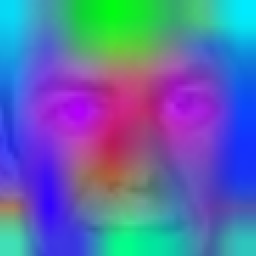} &
    \includegraphics[width=\wifig, valign=c]{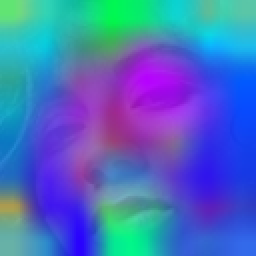} &
    \includegraphics[width=\wifig, valign=c]{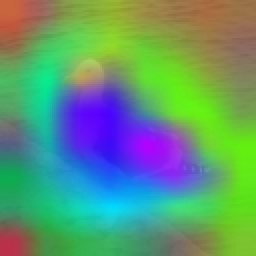} &
    \includegraphics[width=\wifig, valign=c]{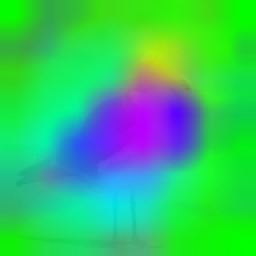} &
    \includegraphics[width=\wifig, valign=c]{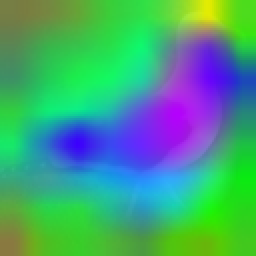} &
    \includegraphics[width=\wifig, valign=c]{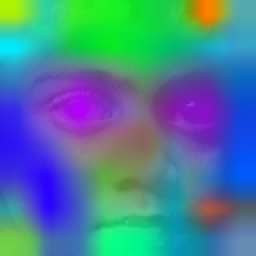} &
    \includegraphics[width=\wifig, valign=c]{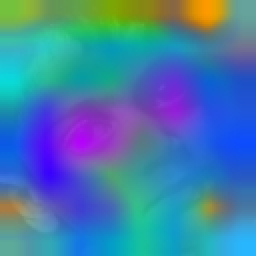} &
    \includegraphics[width=\wifig, valign=c]{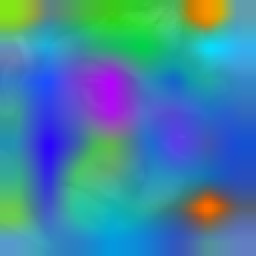} \\
    \end{tabular}
    \caption{\textbf{Semantic parts distillation.} The object parts distilled from our representation using NMF are semantically meaningful and consistent across different instances (left). The parts are also robust to geometric transformations (right).}
    \label{fig:NMF_parts}
    \vspace{-10pt} 
\end{figure}

\noindent
\textbf{Are the learned representations semantically meaningful?}  We found that parts can be reliably distilled from the learned representation using non-negative matrix factorization (NMF) (see~\cite{collins2018deep} for another application of NMF for visualizing semantic parts from deep network activations). 
Fig.~\ref{fig:NMF_parts} shows two such components and a ``map'' of several components which are indicative of parts (left) and are robust to image transformations (right).
Additionally, Fig.~\ref{fig:matching} shows that the correspondence obtained using nearest neighbor matching are semantically meaningful.
Furthermore, our method can be naturally extended to figure-ground segmentation. In appendix we show that the proposed model outperforms an ImageNet pretrained model in few-shot settings (\eg given only 10 annotated images).

\noindent
\textbf{Commonalities and differences between equivariant and invariant learning.}
Equivariance is necessary but not sufficient for an effective landmark representation. It also needs to be distinctive or invariant to nuisance factors.
This is enforced in the equivariance objective (Eqn.~\ref{eq:equivariance2}) as a contrastive term over locations within the same image, as the loss is minimized when $p(v|u;\Phi, \mathbf{x}, \mathbf{x}')$ is maximized at $v=gu$.
This encourages intra-image invariance, unlike the objective of contrastive learning (Eqn.~\ref{eq:infonce}) which encourages inter-image invariance.
However, a single image may contain enough variety to guarantee some invariance. 
This is supported by its empirical performance and recent work showing that representation learning is possible even from a single image~\cite{asano2020a}.
However, our experiments suggest that on challenging datasets, with more clutter, occlusion, and pose variations, inter-image invariance can be more effective.

\noindent
\textbf{Is there any advantage of one approach over the other?} 
Our experiments show that for a deep network of the same size, invariant representation learning can be just as effective (Tab.~\ref{table: face regression}).
However, invariant learning is conceptually simpler and scales better than equivariance approaches, as the latter maintains high-resolution feature maps across the hierarchy. 
Using a deeper network (\eg, ResNet50 vs. ResNet18) gives consistent improvements, outperforming DVE~\cite{DVE} on four out of five datasets, as shown in Tab.~\ref{table: face regression}.
A drawback of our approach is that the hypercolumn representation is not directly interpretable or compact, which results in lower performance in the extreme few-shot case.
However, as seen in Fig.~\ref{fig:curves}a, the advantage disappears with as few as 50 training examples on the AFLW benchmark. This problem can be effectively alleviated by learning a compact representation using equivariant learning which further reduces the number of required training examples to 20.
Invariant learning is also more data-efficient and can achieve the same performance with half the unlabeled examples, as seen in Fig.~\ref{fig:curves}c.
\section{Conclusion}\label{Conclusion}
We show that representations extracted from intermediate layers of a deep network trained using instance-discriminative contrastive learning outperform existing unsupervised landmark representation learning approaches that are based on equivariant learning alone.
We also show these equivariant learning approaches can be viewed through the lens of a (spatial) contrastive learning --- they learn intra-image invariances which can result in weaker generalization than 
inter-image invariances.
Moreover, these two forms of contrastive learning are complementary.
We use the latter to learn a compact representation which leads to a better performance on matching tasks. 
However, by combining these during training can lead to better results.
We illustrate these on existing benchmarks and a new challenging setting where there is a larger variation in pose, viewpoint, and clutter where the improvements are more pronounced. 
We hope that evaluation in the challenging benchmark will encourage novel approaches for learning landmark and part representations.

\section*{Acknowledgements}
  The project is supported in part by Grants \#1661259 and \#1749833
  from the National Science Foundation of United States.
  Our experiments were performed on the University of Massachusetts, Amherst GPU cluster obtained under the Collaborative Fund managed by the
  Massachusetts Technology Collaborative.
  We would also like to thank Erik Learned-Miller, Daniel Sheldon, Rui
  Wang, Huaizu Jiang, Gopal Sharma and Zitian Chen for discussion and feedback on the draft.


{\small
\bibliographystyle{ieee_fullname}
\bibliography{reference}
}

\onecolumn
\appendix
\section*{Appendix}~\label{appendix}
\renewcommand{\thesection}{\Alph{section}}

\noindent The following describes the content in each section in the appendix.
\begin{compactitem}
    \item \S~\ref{sec:others} provides further analysis of the proposed model.
\item \S~\ref{sec:more_matching_regression} shows additional results of landmark matching and regression. 
\item \S~\ref{sec:more_details} describes additional implementation details. 
\item \S~\ref{sec:bird-dataset} details the proposed birds benchmark. 
\item \S~\ref{sec:segmentation} demonstrates the effectiveness of the proposed method for the task of figure-ground segmentation.
\item \S~\ref{sec:number-fig4} provides the numbers corresponding to Figure~\ref{fig:curves}. 
\end{compactitem}

\vspace{0.2in}

\section{Further analysis}~\label{sec:others}
\vspace{-0.2in}
\subsection{Visualization of feature embeddings}~\label{sec:PCA}
\begin{figure} [ht]
    \setlength{\tabcolsep}{0.5pt}
    \newcommand\wifig{0.1\linewidth}
    \centering
    \begin{tabular}{c @{\hspace{3pt}} | @{\hspace{3pt}} cccc @{\hspace{3pt}} | @{\hspace{3pt}} cccc @{\hspace{3pt}}}
    \includegraphics[width=\wifig]{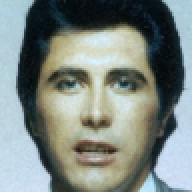} &
    \includegraphics[width=\wifig]{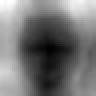} &
    \includegraphics[width=\wifig]{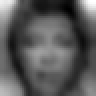} &
    \includegraphics[width=\wifig]{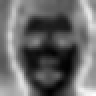} &
    \includegraphics[width=\wifig]{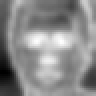} & 
    \includegraphics[width=\wifig]{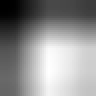} &
    \includegraphics[width=\wifig]{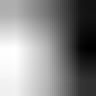} &
    \includegraphics[width=\wifig]{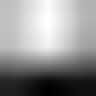} &
    \includegraphics[width=\wifig]{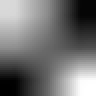} \\
    \includegraphics[width=\wifig]{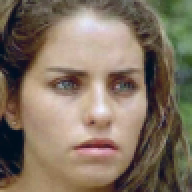} &
    \includegraphics[width=\wifig]{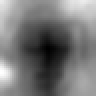} &
    \includegraphics[width=\wifig]{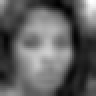} &
    \includegraphics[width=\wifig]{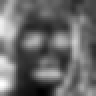} &
    \includegraphics[width=\wifig]{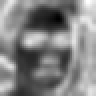} &
    \includegraphics[width=\wifig]{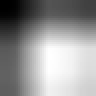} &
    \includegraphics[width=\wifig]{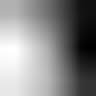} &
    \includegraphics[width=\wifig]{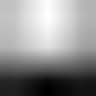} &
    \includegraphics[width=\wifig]{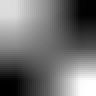} \\
    \multicolumn{1}{c}{Input} & \multicolumn{4}{c}{Contrastively trained model} & \multicolumn{4}{c}{Randomly initialized model}\\
    \end{tabular}
    \caption{\textbf{PCA visualization of the hypercolumn
      representation.} From left to right: input image, and the projection
      of hypercolumns on the first four PCA bases from a contrastively
      trained and a randomly initialized ResNet50.}
    \label{fig:PCA}
\end{figure}

We visualize the first few PCA (uncentered) components of the learned model and a randomly initialized model in Fig.~\ref{fig:PCA}.
Specifically, we sample hypercolumns from 32 MAFL images using our contrastively pre-trained ResNet50, treat each spatial location separately, and compute the PCA basis vectors.
We then project the hypercolumns to each basis and visualize the projection as a spatial map. 
Observe that the bases encode information about the background, foreground, and landmark regions (\eg eyes, nose, and mouth) of faces.
On the other hand, the bases of a randomly initialized model show no such structure.

\subsection{Fine-tuning the network~\label{sec:finetune}}
Tab.~\ref{table: fine-tune} shows the effect of end-to-end fine-tuning of all layers of our model for landmark regression. 
In this experiment, we use the output of the fourth convolutional block of ResNet50 as the representation, which is the optimal single layer representation.
We did \emph{not} find fine-tuning to be uniformly beneficial --- fine-tuning is worse than training the linear regressor only on MAFL dataset, while it is better on AFLW dataset. We speculate the reason to be the domain gap between the datasets used for unsupervised learning and supervised learning of the regressor, which is exacerbated by the small training sets.
For example, AFLW has a larger domain gap than CelebA to MAFL, which is also noticed in DVE~\cite{DVE}.

\begin{table}[!htbp]
  \label{table: fine-tune}
  \centering
  \begin{tabular}{c c c c c c}
    \toprule
                      & MAFL    & AFLW$_M$  & AFLW$_R$ & 300W \\
    \midrule
    w/o fine-tuning    & 2.73   & 8.83     & 7.37    & 6.01 \\
    w/ fine-tuning  & 2.81   & 7.80     & 6.99    & 5.94 \\
    \bottomrule
  \end{tabular}
  \caption{\textbf{Effect of fine-tuning for landmark regression.}
    The fourth block of a ResNet50 network was used as the representation. }
\end{table}

\subsection{Memory efficiency}~\label{sec:memory}
Tab.~\ref{table: efficiency} compares the memory efficiency of DVE~\cite{DVE} to ours. 
DVE maintains high-resolution feature maps across the network hierarchy to compute the equivariance loss. 
By comparison, our contrastive learning loss is computed on a global image representation which requires less memory, allowing bigger models.
Our method with a ``ResNet50-half'', which halves the layer width of the original network, achieves comparable performance with DVE (see Tab.~\ref{tab:matching_appendix} and Tab.~\ref{tab:regression_appendix}) but is more memory-efficient.

\begin{table}[!htbp]
  \label{table: efficiency}
  \centering
  \begin{tabular}{c c c c c}
    \toprule
    Method  & Network  & \# Params (M)  &  Network Size (MB) & Memory (MB) \\
    \midrule
    DVE~\cite{DVE}    & Hourglass  & 12.61  & 48.10 & 491.85 \\
    \midrule
    \multirow{3}{*}{Ours}   & ResNet18   & 11.24 & 42.89  & 11.54\\
                            & ResNet50-half   & 6.03 & 23.02  &  28.15\\
                            & ResNet50   & 23.77 & 90.68  & 52.65\\
    \bottomrule
  \end{tabular}
  \caption{\textbf{Memory efficiency.} Comparison of
    DVE~\cite{DVE} with ours in terms of number of network parameters (\#
    Params), memory required for storing the network (Network Size), and the memory usage of a
    forward and backward pass for a single $96 \times 96$ RGB image
    (Memory).}
\end{table}

\subsection{Effect of dimensionality reduction} 
\begin{figure}[ht!]
    \centering
    \begin{tabular}{c}
    \includegraphics[width=0.4\linewidth]{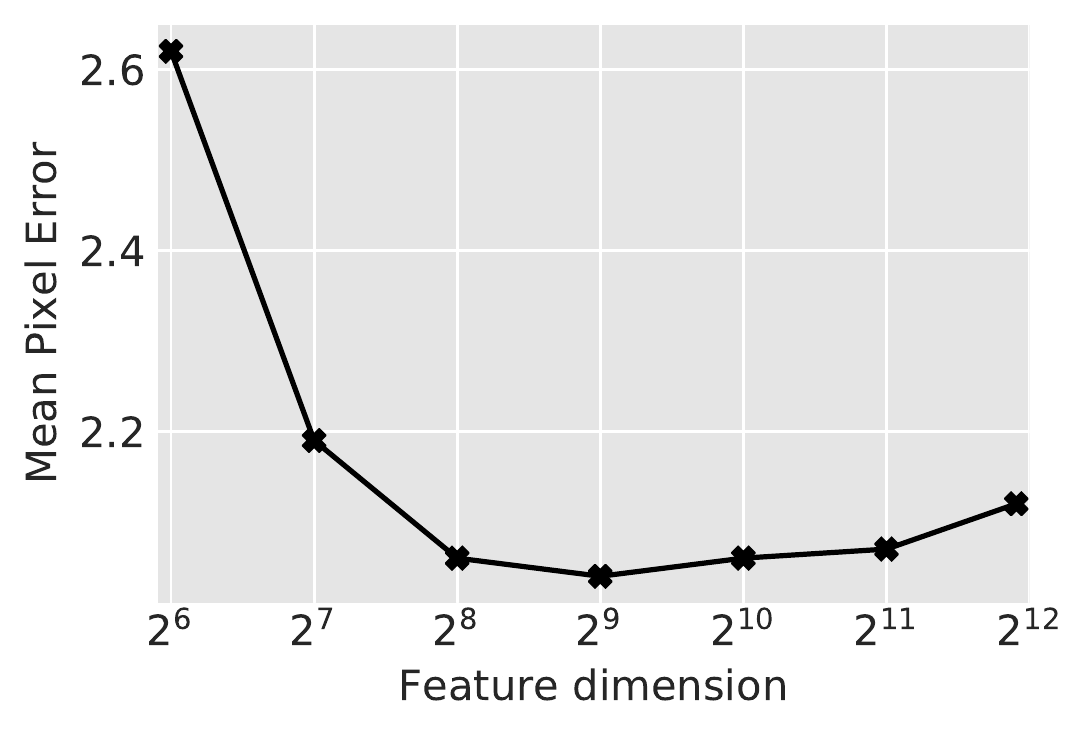} \\
    \end{tabular}
    \caption{\textbf{Landmark matching performance as a function of the projection dimension.} The mean pixel error of the raw hypercolumn representations is 6.16 (not shown), which is higher than the projected representation across all dimensions. }
    \label{fig:intra-learning}
    \vspace{-5pt}
\end{figure}

Fig.~\ref{fig:intra-learning} presents the landmark matching performance as a function of the projection dimension. Specifically, We evaluate the cross-identity landmark matching on the MAFL test set with different projection dimensions. This usually improves performance across all dimensions as the mean pixel error with hypercolumn representations is 6.16.

\subsection{Higher resolution images} \label{sec:high-resolution}
Tab.~\ref{table:high-reso} presents the performance of our model on higher-resolution images. 
Specially, we train our model on $128 \times 128$ CelebA and iNaturalist Aves images (instead of $96 \times 96$ used in the experiments and DVE~\cite{DVE}), 
We conduct the linear evaluation with $128 \times 128$ images from face and bird benchmarks. 
However, the memory and the computational requirement is a challenge where our approach provides an advantage.
A forward and backward pass on a single $128 \times 128$ image takes 874.41 MB for DVE with Hourglass network while it takes only 93.58 MB for the proposed model with ResNet50 as the backbone.
While our method could be trained in 3 days on $128 \times 128$ CelebA images, we could not finish training DVE with Hourglass or ResNet50 network in two weeks despite our best efforts.
We notice that using higher resolution images generally improves the performance across benchmarks.

\begin{table}[ht!]
  \setlength{\tabcolsep}{8pt}
  \renewcommand{\arraystretch}{1.1}
  \centering
  \begin{tabular}{c c c c c c c c}
        \toprule
         Resolution & MAFL & AFLW$_M$ & AFLW$_R$ & 300W  & CUB\\
         \midrule
        $96\times 96$ & 2.44 & 6.99 & \textbf{6.27} & 5.22 & 68.63\\
        $128\times 128$   & \textbf{2.34} & \textbf{6.87} & 6.41 & \textbf{4.99} & \textbf{72.61}\\
        \bottomrule
   \end{tabular}
     \caption{\label{table:high-reso} \textbf{The effect of image resolution.} We use ResNet50 with hypercolumn representations. We report the error in the percentage of inter-ocular distance on human face dataset (\emph{lower is better}), and the percentage of correct keypoints on CUB dataset (\emph{higher is better}).}
\end{table}

\section{Additional results of landmark matching and regression}\label{sec:more_matching_regression}

We provide more experimental results of landmark matching and regression in Tab.~\ref{tab:matching_appendix} and Tab.~\ref{tab:regression_appendix} respectively. We report results under various settings of (1) initialization methods (\eg, randomly initialized, ImageNet pre-trained, DVE~\cite{DVE}, or the proposed contrastively pre-trained); (2) network architectures (\eg, ResNet18, ResNet50, or ResNet50-half which halves the layer width of ResNet50); (3) training dataset for representation learning (aligned or in-the-wild CelebA dataset); (4) representations (hypercolumns or its projected features (+Proj.)). The proposed method surpasses DVE across different network architectures, projected feature dimensions, and curated or uncurated datasets in both landmark matching and regression tasks.

\begin{table}[ht!]
\setlength{\tabcolsep}{7pt}
    \centering
    \begin{tabular}{c c c c c c c c }
        \toprule
        Method & Network & \# Params (M) & In-the-wild  & +Proj. & Dim. & Same identity & Diff. identity\\
        \midrule
        Random & ResNet50 & 23.77 & & & 3840 & 1.07 & 10.03 \\
        Random & ResNet50 & 23.77 & & $\checkmark$& 256 & 3.68 & 7.04 \\
        Random & ResNet50 & 23.77 & & $\checkmark$& 128 & 3.70 & 7.03 \\
        Random & ResNet50 & 23.77 & & $\checkmark$& 64 & 3.71 & 7.10 \\
        \midrule
        ImageNet & ResNet50  &  23.77 & & & 3840 & 0.67 & 6.50 \\
        ImageNet & ResNet50  &  23.77 & & $\checkmark$& 256 & 0.82 & 3.15 \\
        ImageNet & ResNet50  &  23.77 & & $\checkmark$& 128 & 1.00 & 3.39 \\
        ImageNet & ResNet50  &  23.77 & & $\checkmark$& 64 & 1.55 & 4.44 \\
        \midrule
        DVE~\cite{DVE} & Smallnet &  0.35 &  &  & 64  & 1.28 & 2.77 \\
        DVE~\cite{DVE} & Hourglass & 12.61 &  &  & 64  & 0.92 & 2.38 \\
        DVE~\cite{DVE} & Hourglass & 12.61 & $\checkmark$ &  & 64 & 1.27 & 3.52 \\
        \midrule
        Ours & ResNet50 & 23.77 & &  & 3840 & \textbf{0.73} & 6.16 \\
        Ours & ResNet50 & 23.77 & & $\checkmark$& 256 & \textbf{0.71} & \textbf{2.06} \\
        Ours & ResNet50 & 23.77 & & $\checkmark$& 128 & \textbf{0.82} & \textbf{2.19} \\
        Ours & ResNet50 & 23.77 & & $\checkmark$& 64 & \textbf{0.92} & 2.62 \\
        \midrule
        Ours & ResNet50 & 23.77 & $\checkmark$ & & 3840 & \textbf{0.78} & 5.58 \\
        Ours & ResNet50 & 23.77 & $\checkmark$ &  $\checkmark$& 256 & \textbf{0.96} & \textbf{3.03} \\
        Ours & ResNet50 & 23.77 & $\checkmark$ &  $\checkmark$& 128 & \textbf{0.98} & \textbf{3.05} \\
        Ours & ResNet50 & 23.77 & $\checkmark$ & $\checkmark$ & 64  & \textbf{0.99} & \textbf{3.06} \\
        \midrule
        Ours & ResNet50-half & 6.03 & & & 3840 & \textbf{0.74} & 5.84 \\
        Ours & ResNet50-half & 6.03 &  &  $\checkmark$& 256 & \textbf{0.76} & \textbf{2.18} \\
        Ours & ResNet50-half & 6.03 &  &  $\checkmark$& 128 & \textbf{0.88} & \textbf{2.38} \\
        Ours & ResNet50-half & 6.03 &  &  $\checkmark$& 64 & 1.05 & 2.85 \\
        \midrule
        Ours & ResNet18 & 11.24 & &  & 3840 & \textbf{0.64} & 4.95 \\
        Ours & ResNet18 & 11.24 & &  $\checkmark$& 256 & \textbf{0.71} & \textbf{2.20} \\
        Ours & ResNet18 & 11.24 & &  $\checkmark$& 128 & \textbf{0.82} & \textbf{2.31} \\
        Ours & ResNet18 & 11.24 & &  $\checkmark$ & 64  & 1.00 & 2.74 \\
        \bottomrule
    \end{tabular}
    
    \caption{\textbf{Landmark matching}. The mean pixel error between the predicted landmarks and the ground-truth (\emph{lower is better}). Results better than DVE's are in \textbf{bold}.}
    \label{tab:matching_appendix}
\end{table}

\begin{table}[ht!]
    \centering
    \begin{tabular}{c c c c c c c c c c}
        \toprule
        Method & Network & \# Params (M) & In-the-wild  & +Proj. & Dim. &  MAFL & AFLW$_M$ & AFLW$_R$ & 300W \\
        \midrule
        Random & ResNet50 & 23.77 & &  & 3840  & 4.72 & 16.74 & 11.23 & 11.70\\
        Random & ResNet50 & 23.77 & & $\checkmark$& 256  & 6.56 & 20.33 & 13.50 & 11.67 \\
        Random & ResNet50 & 23.77 & & $\checkmark$& 128  & 6.21 & 20.25 & 13.99 & 17.12 \\
        Random & ResNet50 & 23.77 & & $\checkmark$& 64   & 6.28 & 19.76 & 13.53 & 16.93 \\
        \midrule
        ImageNet & ResNet50  & 23.77 & &  & 3840  & 2.98 & 8.88 & 7.34 & 6.88\\
        ImageNet & ResNet50  & 23.77 & & $\checkmark$& 256 & 3.51 & 9.69 & 8.02 & 7.02 \\
        ImageNet & ResNet50  & 23.77 & & $\checkmark$& 128  & 3.36 & 9.11 & 7.68 & 6.55 \\
        ImageNet & ResNet50  & 23.77 & & $\checkmark$& 64  & 3.50 & 9.43 & 7.63 & 6.51 \\
        \midrule
        DVE & Smallnet  & 0.35  &  &  & 64  & 3.42  & 8.60 & 7.79 & 5.75 \\
        DVE & Hourglass & 12.61 &  &  & 64  & 2.86  & 7.53 & 6.54 & 4.65  \\
        DVE & Hourglass & 12.61 & $\checkmark$ &  & 64  & 3.23 & 8.52 & 7.38 & 5.05 \\
        \midrule
        Ours & ResNet50 & 23.77 &  & & 3840  & \textbf{2.44} & \textbf{6.99} & \textbf{6.27} & 5.22\\
        Ours & ResNet50 & 23.77 && $\checkmark$& 256   & \textbf{2.64} & \textbf{7.17} & \textbf{6.14} & 4.99\\
        Ours & ResNet50 &23.77 & & $\checkmark$& 128   & \textbf{2.71} & \textbf{7.14} & \textbf{6.14} & 5.09 \\
        Ours & ResNet50 &23.77 & & $\checkmark$& 64   & \textbf{2.77} & \textbf{7.21} & \textbf{6.22} & 5.19\\
        \midrule
        Ours & ResNet50 &23.77 & $\checkmark$ & & 3840  & \textbf{2.46} & \textbf{7.57} & \textbf{6.29} & \textbf{5.04} \\
        Ours & ResNet50 &23.77 & $\checkmark$ & $\checkmark$& 256  & \textbf{2.82} & \textbf{7.69} & \textbf{6.67} & 5.27 \\
        Ours & ResNet50 &23.77 & $\checkmark$ & $\checkmark$& 128  & \textbf{2.88} & \textbf{7.81} & \textbf{6.79} & 5.37 \\
        Ours & ResNet50 &23.77 & $\checkmark$ & $\checkmark$& 64   & \textbf{3.00} & \textbf{7.87} & \textbf{6.92} & 5.59 \\
        \midrule
        Ours & ResNet50-half & 6.03 &  &  & 3840  & \textbf{2.46}  & \textbf{7.37} & 6.71 & 5.33 \\
        Ours & ResNet50-half & 6.03 &  & $\checkmark$& 256  & \textbf{2.66} & \textbf{7.21} & \textbf{6.32} & 5.20 \\
        Ours & ResNet50-half & 6.03 &  & $\checkmark$& 128  & \textbf{2.75} & \textbf{7.26} & \textbf{6.32} & 5.26 \\
        Ours & ResNet50-half & 6.03 &  & $\checkmark$& 64   & \textbf{2.85} & \textbf{7.42} & \textbf{6.45} & 5.42 \\
        \midrule
        Ours & ResNet18 & 11.24 &  &  & 3840 & \textbf{2.57} & 8.59 & 7.38 & 5.78 \\
        Ours & ResNet18 & 11.24 & & $\checkmark$ & 256  & \textbf{2.71} & \textbf{7.23} & \textbf{6.30} & 5.20\\
        Ours & ResNet18 & 11.24 & & $\checkmark$ & 128  & \textbf{2.81} & \textbf{7.30} & \textbf{6.32} & 5.30 \\
        Ours & ResNet18 & 11.24 & & $\checkmark$ & 64   & 2.89 & \textbf{7.48} & \textbf{6.43} & 5.42 \\
        \bottomrule
    \end{tabular}
    \caption{\textbf{Landmark regression.} The error in the percentage of inner-ocular distance (\emph{lower is better}). Results better than DVE's are in \textbf{bold}.}
    \label{tab:regression_appendix}
\end{table}

\section{Other implementation details}\label{sec:more_details}

\paragraph{Training details of unsupervised learning models.} We use MoCo \cite{he2019momentum} as our contrastive learning model. We train MoCo for 800 epochs with a batch size of 256 and a cosine learning rate schedule as proposed in MoCo-v2~\cite{chen2020improved}. However, we did not observe improvements in our task when using other tricks in MoCo-v2~\cite{chen2020improved}, such as adding Gaussian blur for the data augmentation and using an MLP as the projection network. We use the public implementation\footnote{\url{https://github.com/HobbitLong/CMC}} of MoCo from~\cite{Tian2019}.
For a comparison with the DVE model~\cite{DVE} on the proposed bird dataset, we use their publicly available implementation\footnote{\url{https://github.com/jamt9000/DVE}}. 

\paragraph{Training details of feature projection.} We set the temperature hyperparameter $\tau$ to $1/7$ in the equivariance loss (Equation 3 in the main text) for training the linear feature projector. We implement the linear projector as a $1 \times 1$ convolutional layer and train the projector for 10 epochs on the CelebA dataset~\cite{liu2015deep} with the equivariance loss. Notice that we do not apply any data augmentations during training, and the training of feature projector does not require any human annotations. We use Adam optimizer with a learning rate of 0.001 and a weight decay of 0.0005.

\paragraph{Training details of linear regression.} 
We do not use any data augmentation when the \emph{entire} annotations of the training data are provided (Tab.~\ref{table: face regression}).
However, in the \emph{limited} annotation experiments on human face benchmarks, following DVE~\cite{DVE}, we apply thin-plate spline as the data augmentation method with the same deformation hyperparameters as DVE (Fig.~\ref{fig:curves}a). 
We do not apply any augmentations during landmark regression on the CUB dataset (Tab.~\ref{table: face regression} and Fig.~\ref{fig:curves}b). 
Due to the lack of validation set on face benchmarks, we train the linear regressor for 120, 45, and 80 epochs on MAFL, AFLW, and 300W dataset respectively when hypercolumns are used, and we train for 150 epochs uniformly across these benchmarks when the compact representations of the hypercolumn are used. 
On CUB, the results are reported from the checkpoint selected on the validation set. 
On face benchmarks, we use an initial learning rate of 0.01 and a weight decay of 0.05 when only limited annotations are available (Fig.~\ref{fig:curves}a,b). The two hyperparameters are 0.001 and 0.0005 respectively when the entire annotations are given (Tab.~\ref{table: face regression}).
On CUB, when the number of annotations is less than 100, we use an initial learning rate of 0.01 and a weight decay of 0.05 for both ResNet and Hourglass network. When more annotations are available, the learning rate is 0.01 for both ResNet and Hourglass network, while the weight decay is 0.005 for ResNet50 and 0.0005 for Hourglass. These hyperparameters are selected on the evaluation set.
In the ablation study of the effectiveness of unsupervised learning (Tab.~\ref{table: random net}), for the ImageNet pre-trained or randomly initialized networks, we report the best performance on the test set within 2000 training epochs. 
We apply the cosine learning rate schedule in all of our experiments.

\section{Birds benchmark~\label{sec:bird-dataset}}
\begin{figure} [ht!]
    \setlength{\tabcolsep}{0.8pt}
    \newcommand\wifig{0.1\linewidth}
    \centering
    \begin{tabular}{cccccccccc}
    \rotatebox[origin=c]{90}{iNat Aves} &
   \includegraphics[width=\wifig, valign=c]{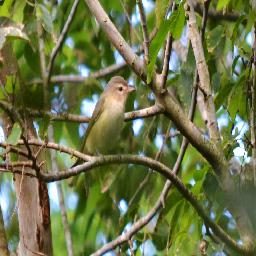} &
    \includegraphics[width=\wifig, valign=c]{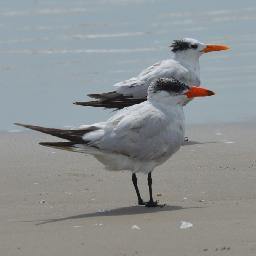} &
    \includegraphics[width=\wifig, valign=c]{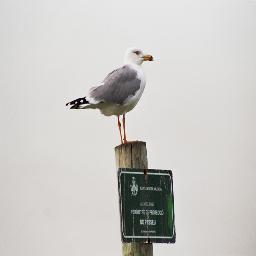} &
    \includegraphics[width=\wifig, valign=c]{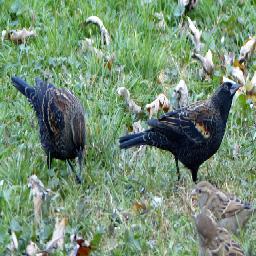} &
    \includegraphics[width=\wifig, valign=c]{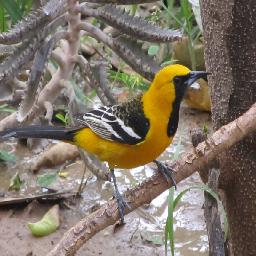} &
    \includegraphics[width=\wifig, valign=c]{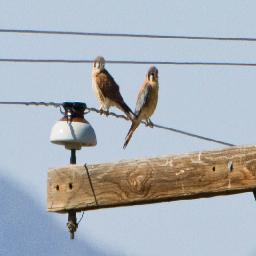} &
    \includegraphics[width=\wifig, valign=c]{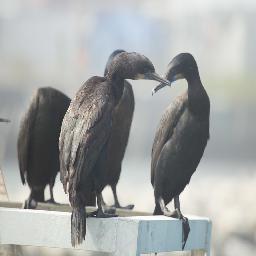} &
    \includegraphics[width=\wifig, valign=c]{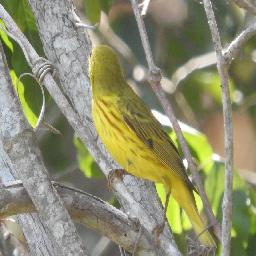} &
    \includegraphics[width=\wifig, valign=c]{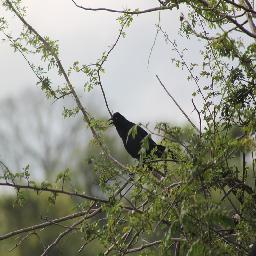} \\ 
    \vspace{-0.1in}
    \\
    \rotatebox[origin=c]{90}{CUB} &
   \includegraphics[width=\wifig, valign=c]{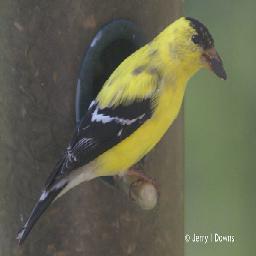} &
    \includegraphics[width=\wifig, valign=c]{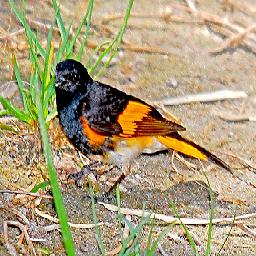} &
    \includegraphics[width=\wifig, valign=c]{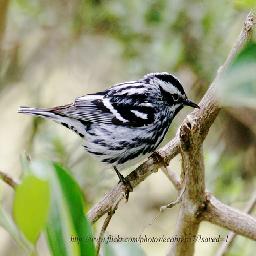} &
    \includegraphics[width=\wifig, valign=c]{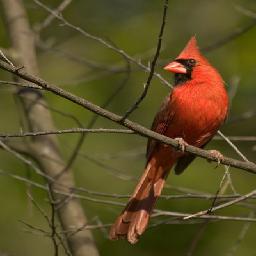} &
    \includegraphics[width=\wifig, valign=c]{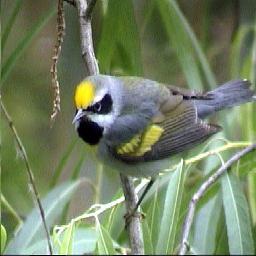} &
    \includegraphics[width=\wifig, valign=c]{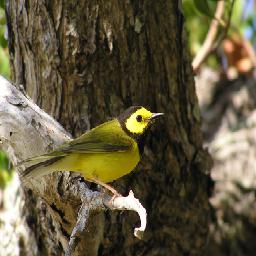} &
    \includegraphics[width=\wifig, valign=c]{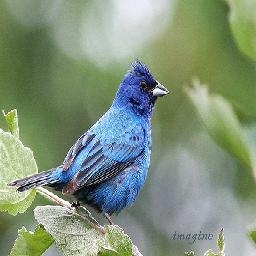} &
    \includegraphics[width=\wifig, valign=c]{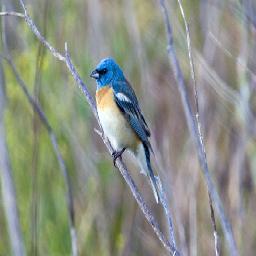}&
    \includegraphics[width=\wifig, valign=c]{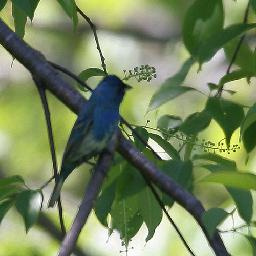}  \\
    \end{tabular}
    \caption{\textbf{Images from bird datasets.} Images in the CUB dataset (bottom)
      are iconic with birds more frequently in canonical poses and
      contain a single instance. On the other hand, iNaturalist images (top)
      are community driven and less curated. Often multiple birds are
      in a single image and are far away. This makes learning and
      transfer more challenging.}
    \label{fig:inat_cub}
\end{figure}

The Birds benchmark consists of unsupervised learning on the images from the iNaturalist Aves taxa and evaluating them on the landmarks in the CUB dataset. 
Specially, we randomly sample 100K images of birds from the iNaturalist 2017 dataset~\cite{gvanhorn2018inat} under ``Aves'' class. 
Fig.~\ref{fig:inat_cub} top row shows images from iNaturalist Aves dataset.
The dataset contains objects in significant clutter, occlusion, and with a wider range of pose, viewpoint, and size variations than those in face benchmarks.
Some images even contain multiple objects. 
To test the performance in the few-shot setting, we sample a subset of the CUB dataset which contains similar species to iNaturalist. 
Specifically, we sample 35 species of \textit{Passeroidea} superfamily, each annotated with 15 landmarks. 
Fig.~\ref{fig:inat_cub} bottom row shows images from the CUB dataset.
We sample at most 60 images per class and conduct the splitting of training, validation, and test set on the samples of each species in a ratio of 3:1:1. These splits are then combined, which results in 1241 training images, 382 validation images, and 383 test images.

\section{Figure-ground Segmentation} \label{sec:segmentation}
We extend the model to tackle a figure-ground segmentation task.
Specifically, we train a 1$\times$1 convolutional layer on the top of the learned pixel representations to predict the foreground mask. 
We evaluate the segmentation model on the CUB dataset described in Sec.~\ref{sec:bird-dataset}, which comes with 
annotated masks.
Tab.~\ref{table:segmentation} compares representations from randomly initialized, ImageNet pre-trained, and our contrastively learned networks under different sizes of the training set. 
The contrastive model is trained on iNaturalist Aves dataset, as described in the main text. Under the linear evaluation setting where the backbone is fixed, the contrastive model with hypercolumn representation outperforms both the randomly initialized and ImageNet pre-trained models. We are unable to get meaningful results for the ImageNet pre-trained model with fewer than 100 annotations and for a randomly initialized network with the entire dataset.
Fine-tuning the network end-to-end outperforms training only the linear layer across different representations.
Hypercolumns are more effective than the activations from the fourth convolutional block with fine-tuning and linear evaluation (Tab.~\ref{table:segmentation}).

\begin{figure*} [ht]
    \setlength{\tabcolsep}{1pt}
    \centering
    \begin{tabular}{ccccccc}
    \rotatebox[origin=c]{90}{Ground Truth} &
    \includegraphics[width=0.12\linewidth, valign=c]{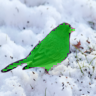} &
    \includegraphics[width=0.12\linewidth, valign=c]{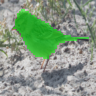} &
    \includegraphics[width=0.12\linewidth, valign=c]{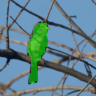} &
    \includegraphics[width=0.12\linewidth, valign=c]{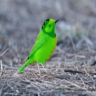} &
    \includegraphics[width=0.12\linewidth, valign=c]{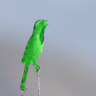} &
    \includegraphics[width=0.12\linewidth, valign=c]{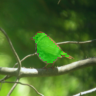} \\
    \rotatebox[origin=c]{90}{Random} &
    \includegraphics[width=0.12\linewidth, valign=c]{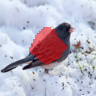} &
    \includegraphics[width=0.12\linewidth, valign=c]{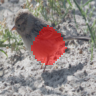} &
    \includegraphics[width=0.12\linewidth, valign=c]{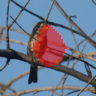} &
    \includegraphics[width=0.12\linewidth, valign=c]{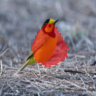} &
    \includegraphics[width=0.12\linewidth, valign=c]{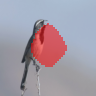} &
    \includegraphics[width=0.12\linewidth, valign=c]{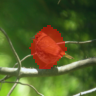} \\
    \rotatebox[origin=c]{90}{Contrastive} &
    \includegraphics[width=0.12\linewidth, valign=c]{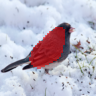} &
    \includegraphics[width=0.12\linewidth, valign=c]{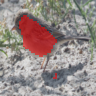} &
    \includegraphics[width=0.12\linewidth, valign=c]{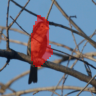} &
    \includegraphics[width=0.12\linewidth, valign=c]{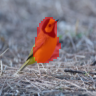} &
    \includegraphics[width=0.12\linewidth, valign=c]{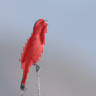} &
    \includegraphics[width=0.12\linewidth, valign=c]{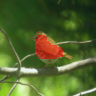} \\
    \end{tabular}
    \caption{\textbf{Figure-ground segmentation on CUB dataset with 10 annotated images as training data.} We fine-tune the network end-to-end using the hypercolumn representation.}
      \label{fig:segmentation}
\end{figure*}

One observation is fine-tuning a randomly initialized network achieves good quantitative performance on this task. A closer inspection, as presented in Figure~\ref{fig:segmentation}, reveals that this is because the randomly initialized network simply generates a fixed mask at the center of each test which results in high intersection-over-union with the object mask as most images from the CUB dataset are object-centric (as shown in Fig.~\ref{fig:inat_cub}). Evaluation with a boundary-metric might reveal this difference. Our model on the other hand achieves meaningful and highly accurate masks with as few as 10 training images, as shown in Fig.~\ref{fig:segmentation}.

\begin{table}[!htbp]
  \centering
  \begin{tabular}{c c c c c c c c c}
    \toprule
    \multirow{2}{*}{Self-supervision} & Backbone & \multirow{2}{*}{Hypercolumn} & \multicolumn{5}{c}{\# of annotation}\\
                 &   Fixed?   & &10       & 50       & 100       & 250    &1241  \\
    \midrule
    Random       &  \checkmark  &  \checkmark   & 0.00     & 0.14    & 0.01      & 0.00   & 0.12  \\
    ImageNet     &  \checkmark  &  \checkmark   & 0.12      & 0.08    & 0.22      & 0.51     & \textbf{0.66} \\
    Contrastive  &  \checkmark  &  \checkmark   & 0.36      & \textbf{0.52}    & \textbf{0.59}     & \textbf{0.61}  & 0.62 \\
    Contrastive  &  \checkmark  &  $\times$   &  \textbf{0.37}     & 0.45    &  0.51    &   0.52    &  0.53\\
    \midrule
    Random       &  $\times$    &  \checkmark   & 0.41     & 0.40    & 0.48      & 0.55   & 0.71  \\
    ImageNet     &  $\times$    &  \checkmark    &  0.38     &  \textbf{0.56}   &    \textbf{0.58}   &  \textbf{0.63}    &  0.73 \\
    Contrastive  &  $\times$    &  \checkmark   & \textbf{0.46}      & 0.48    & 0.54     & \textbf{0.63}      & \textbf{0.74} \\
    Contrastive  &  $\times$    &  $\times$   & 0.39      & 0.43    & 0.45     & 0.51      & 0.59 \\
    \bottomrule
  \end{tabular}
    \caption{\textbf{Figure-ground segmentation on CUB dataset.} We report the mean Intersection-over-Union (IoU) performance (\emph{higher is better}) using a ResNet50 network.}
  \label{table:segmentation}
\end{table}

\section{Tables for Figure~\ref{fig:curves}\label{sec:number-fig4}}

Tab.~\ref{table: limited annotation},~\ref{table:CUB_limit_anno}, and \ref{table:trainset} present the numbers corresponding to Fig.~\ref{fig:curves}a, b, and c respectively. 
These describe the effect of dataset size for landmark regression and unsupervised learning.

\vspace{0.2in}

\begin{table}[!htbp]
  \setlength{\tabcolsep}{8pt}
  \renewcommand{\arraystretch}{1.1}
  \centering
  \begin{tabular}{l c c c c c c c}
    \toprule
    \multirow{2}{*}{Self-supervision}& \multicolumn{7}{c}{\# of annotations} \\
       &1  &5    &10   & 20      & 50   & 100   & 10122\\
    \midrule
    
    None (SmallNet) \cite{DVE}   & 28.87    & 32.85     & 22.31       & 21.13     &    --        &    --    & 14.25   \\
    
    \hline
    \multirow{2}{*}{DVE (Hourglass)~\cite{DVE}}  & 14.23  &  12.04    &12.25   & 11.46    &  12.76     &     11.88  &  7.53\\
                & \scriptsize{$\pm$1.54} & \scriptsize{$\pm$2.03} & \scriptsize{$\pm$2.42} & \scriptsize{$\pm$0.83} & \scriptsize{$\pm$0.53} &  \scriptsize{$\pm$ 0.16}  \\
    
    \hline
    \multirow{2}{*}{Ours (ResNet50 + hypercol.)}        &   42.69    &    25.74    &  17.61     &  13.35     &  10.67     &   9.24    &   6.99   \\
                & \scriptsize{$\pm$5.10} & \scriptsize{$\pm$2.33} & \scriptsize{$\pm$0.75} & \scriptsize{$\pm$0.33} & \scriptsize{$\pm$0.35} &  \scriptsize{$\pm$ 0.35} \\
    \multirow{2}{*}{Ours (ResNet50 + conv4)}        &  43.74 & 21.25 & 16.51 & 12.45 & 10.03 & 9.95 & 8.05   \\
                & \scriptsize{$\pm$ 2.78} & \scriptsize{$\pm$1.14} & \scriptsize{$\pm$1.43} & \scriptsize{$\pm$0.66} & \scriptsize{$\pm$0.21} &  \scriptsize{$\pm$ 0.17} \\
    
    \multirow{2}{*}{Ours (ResNet50 + 256D proj.)}        &  28.00  &  15.85  &  12.98  & 11.18  &  9.56  &  9.30    &  7.17   \\
                & \scriptsize{$\pm$1.39} & \scriptsize{$\pm$0.86} & \scriptsize{$\pm$0.16} & \scriptsize{$\pm$0.19} & \scriptsize{$\pm$0.44} &  \scriptsize{$\pm$ 0.20} \\
    \multirow{2}{*}{Ours (ResNet50 + 128D proj.)}      & 27.31  & 18.66 & 13.39  & 11.77 & 10.25 & 9.46 & 7.14  \\
                & \scriptsize{$\pm$ 1.39} & \scriptsize{$\pm$4.59} & \scriptsize{$\pm$0.30} & \scriptsize{$\pm$0.85} & \scriptsize{$\pm$0.22} &  \scriptsize{$\pm$ 0.05} \\
    \multirow{2}{*}{Ours (ResNet50 + 64D proj.)}      & 24.87  & 15.15 & 13.62  & 11.77 & 11.57 & 10.06 & 7.21  \\
                & \scriptsize{$\pm$ 2.67} & \scriptsize{$\pm$0.53} & \scriptsize{$\pm$1.08} & \scriptsize{$\pm$0.68} & \scriptsize{$\pm$0.03} &  \scriptsize{$\pm$ 0.45} \\

    \multirow{2}{*}{Ours (ResNet18 + hypercol.)}     & 47.15 & 24.99 & 17.40 & 13.87 & 11.04 & 9.93 & 8.59   \\
                & \scriptsize{$\pm$ 6.88} & \scriptsize{$\pm$3.21} & \scriptsize{$\pm$0.37} & \scriptsize{$\pm$0.66} & \scriptsize{$\pm$0.92} &  \scriptsize{$\pm$0.39 } \\
    \multirow{2}{*}{Ours (ResNet18 + conv4)}        & 38.05 & 21.71 & 16.60 & 14.48 & 12.20 & 11.02 & 10.61  \\
                & \scriptsize{$\pm$5.25} & \scriptsize{$\pm$1.57} & \scriptsize{$\pm$0.61} & \scriptsize{$\pm$0.69} & \scriptsize{$\pm$0.36} &  \scriptsize{$\pm$0.06} \\
    \bottomrule \\
  \end{tabular}
    \caption{\textbf{Landmark regression with limited annotations on
      AFLW$_M$.} The results are reported as the error in percentage of
    inter-ocular distance (\emph{lower is better}).}
  \label{table: limited annotation}
\end{table}

\begin{table}[!htbp]
  \setlength{\tabcolsep}{8pt}
  \renewcommand{\arraystretch}{1.1}
  \centering
  \begin{tabular}{l c c c c c c c}
    \toprule
    \multirow{2}{*}{Self-supervision} & \multicolumn{6}{c}{\# of annotation}\\

                       & 10       & 50       & 100       & 250     & 500    &1241  \\
    \midrule
    None (ResNet18)              & 2.97     &  10.07      &  11.31      &  24.82     & 38.86    & 52.64       \\
    DVE (Hourglass)~\cite{DVE}     & 37.82    &  51.64      &  54.58      &  56.78     &  58.64   & 61.91   \\
    Ours (ResNet18 + hypercol.)    & 13.41    &  25.91      &  34.02      &  51.70     &  56.77   & 62.24     \\
    Ours (ResNet50 + hypercol.)    & 13.87    &  29.28      &  40.86      &  57.96     &  64.55   & 68.63      \\
    Ours (ResNet50 + 256D proj.) &  16.32 & 38.70 & 48.75 & 56.04 & 57.74 & 61.22 \\
    Ours (ResNet50 + 512D proj.) &  17.29 & 43.90 & 49.91 & 57.96 & 58.93 & 62.55 \\
    Ours (ResNet50 + 1280D proj.) &  18.94 & 47.02 & 50.75 & 57.24 & 59.89 & 63.25 \\
    \bottomrule\\
  \end{tabular}
    \caption{\textbf{Landmark regression on bird dataset.} The results
    are reported as percentage of correct keypoints (PCK). (\emph{higher is better}).}
  \label{table:CUB_limit_anno}
\end{table}

\begin{table}[!htbp]
  \setlength{\tabcolsep}{8pt}
  \renewcommand{\arraystretch}{1.1}
  \centering
  \begin{tabular}{c c c c c  c c}
    \toprule
    \multirow{2}{*}{Methods} &  \multirow{2}{*}{Dimension} & \multicolumn{5}{c}{Training set size}  \\
                 & & 5\%       & 10\%       & 25\%       & 50\%     & 100\%    \\
    \midrule
    DVE              & 64   & --      &  --      &  --      &  --   & 7.53 \\
    Ours   & 3840  & 13.26     &  9.12      & 7.82       &  7.22    & 6.99  \\
    Ours+proj.   & 256  & 8.88      &  8.20      & 7.54       & 7.32     & 7.17 \\
    Ours+proj.   & 128  & 9.31      & 8.50      & 7.64        & 7.29    & 7.14\\
    Ours+proj.   & 64   & 9.41      & 9.69      & 8.27       & 7.60     & 7.21\\
    \bottomrule \\
  \end{tabular}
    \caption{\textbf{The effect of training set size on unsupervised
      learning models.} The results are reported as percentage of
    inter-ocular distance on AFLW$_M$ benchmark (\emph{lower is better}).}
  \label{table:trainset}
\end{table}

\end{document}